\documentclass[11pt,letterpaper]{article}
\usepackage[margin=1in]{geometry}
\usepackage{times}
\usepackage[utf8]{inputenc}     \usepackage[T1]{fontenc}        \usepackage{xurl}               \usepackage{booktabs}           \usepackage{amsfonts}           \usepackage{nicefrac}           \usepackage{microtype}          

\usepackage{amsthm}           \usepackage{physics}          \usepackage{graphicx}         \usepackage{wrapfig}          \usepackage{bm}               \usepackage{bbm}              \usepackage{amssymb}          \usepackage{mathtools}        \usepackage{mathrsfs}         \usepackage{siunitx}          \usepackage{tikz, tikz-cd}

\DeclareMathOperator*{\argmax}{arg\,max}

\definecolor{myblue}{cmyk}{1.,.7,0,.3}
\definecolor{myorange}{cmyk}{0,.6,1.,0}

\DeclareFontFamily{U}{wncy}{}
\DeclareFontShape{U}{wncy}{m}{n}{<->wncyr10}{}
\DeclareSymbolFont{mcy}{U}{wncy}{m}{n}
\DeclareMathSymbol{\Sh}{\mathord}{mcy}{"58}

\usepackage{multirow}
\usepackage{tabu}

\usepackage{natbib}
\setcitestyle{authoryear,round,citesep={;},aysep={,},yysep={;}}
\usepackage{hyperref} \usepackage{placeins}
\usetikzlibrary{arrows.meta,positioning,decorations.pathreplacing,calc,patterns,fit}
\definecolor{myblue}{HTML}{1F4E79}
\definecolor{myorange}{HTML}{E08E0B}

\providecommand{\erfc}{\operatorname{erfc}}
\providecommand{\softmax}{\operatorname{softmax}}
\providecommand{\R}{\mathbb{R}}

\newcommand{\papertitle}{FreeSolo: Linear-time Sparse Attention with Zero-Shot Length Extrapolation}
\hypersetup{hidelinks,pdftitle={\papertitle}}

\title{\bfseries\papertitle}
\author{Zekun Shi\textsuperscript{1,3}\qquad Alastair James Arthur Price\textsuperscript{2}\\[0.5em]
  {\small\textsuperscript{1}National University of Singapore}\\
  {\small\textsuperscript{2}University of Toronto\qquad
         \textsuperscript{3}Sea AI Lab, Singapore}\\[0.35em]
  {\small\href{mailto:zekun.shi@comp.nus.edu.sg}{\texttt{zekun.shi@comp.nus.edu.sg}}}}
\date{}
\hypersetup{pdfauthor={Zekun Shi and Alastair James Arthur Price}}
\newcommand{\methodpagebreak}{}
\newcommand{\referencespagebreak}{\FloatBarrier}
\newcommand{\mainresultstablefont}{\fontsize{7.5}{9}\selectfont}

\newcommand{\codereleasefootnote}{We will open-source the code.}

\begin{document}
\maketitle
\begin{abstract}
Linear-time sequence models typically store long-context information in a recurrent state. A fixed-size state limits how much information can be stored, while hybrid architectures that add full attention for better recall incur quadratic sequence cost. Alternatively, sparse attention with continued pretraining (CPT) can extend the effective context length. Generalization beyond the trained length remains challenging, however, and CPT can be costly. We introduce \textsc{FreeSolo}\footnote{\codereleasefootnote}, a native sparse attention mechanism with a short-range sliding window branch and a long-range branch that selects keypoint cells for direct token attention. Its linear-time variant, FS-L, learns to prune the prefix before query-dependent selection through short CPT. FS-B searches balanced trees with $O(N\log^2 N)$ algorithmic cost including radix-sort construction. FreeSolo demonstrates better zero-shot length extrapolation than the baselines without long-context fine-tuning.
\end{abstract}

\begin{figure}[!ht]
\centering
\includegraphics[width=\linewidth]{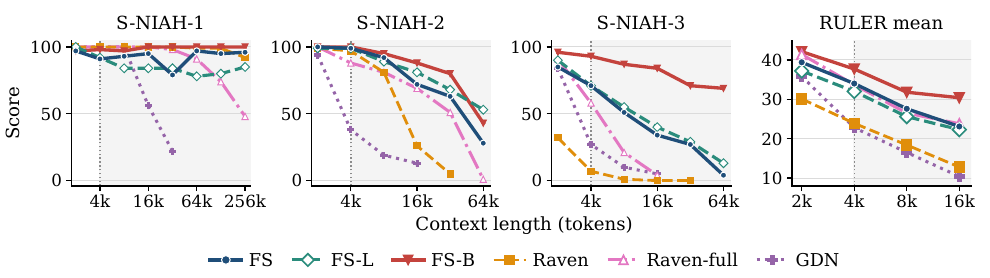}
\caption{\textbf{Zero-shot RULER extrapolation at $800$M.}
Models train at $4$k (dotted line); shading marks extrapolation.
$100$ examples per task and length.}
\label{fig:extrap}
\end{figure}

\section{Introduction}
\label{sec:intro}
Long-context language modeling requires efficient attention and retrieval beyond
the training length. Full attention has quadratic sequence
cost~\citep{vaswani2017attention}. Recurrent models compress history into fixed-size
states~\citep{katharopoulos2020linear,gu2023mamba,dao2024mamba2}, while Native Sparse
Attention (NSA)~\citep{yuan2025nsa} and Mixture of Block Attention
(MoBA)~\citep{lu2025moba} preserve token access but score growing sets of blocks.
Neither compression nor sparsity alone
ensures length extrapolation. Continued pretraining (CPT) on longer sequences
can extend retrieval at additional training cost~\citep{fu2024context}.

\begin{figure}[t]
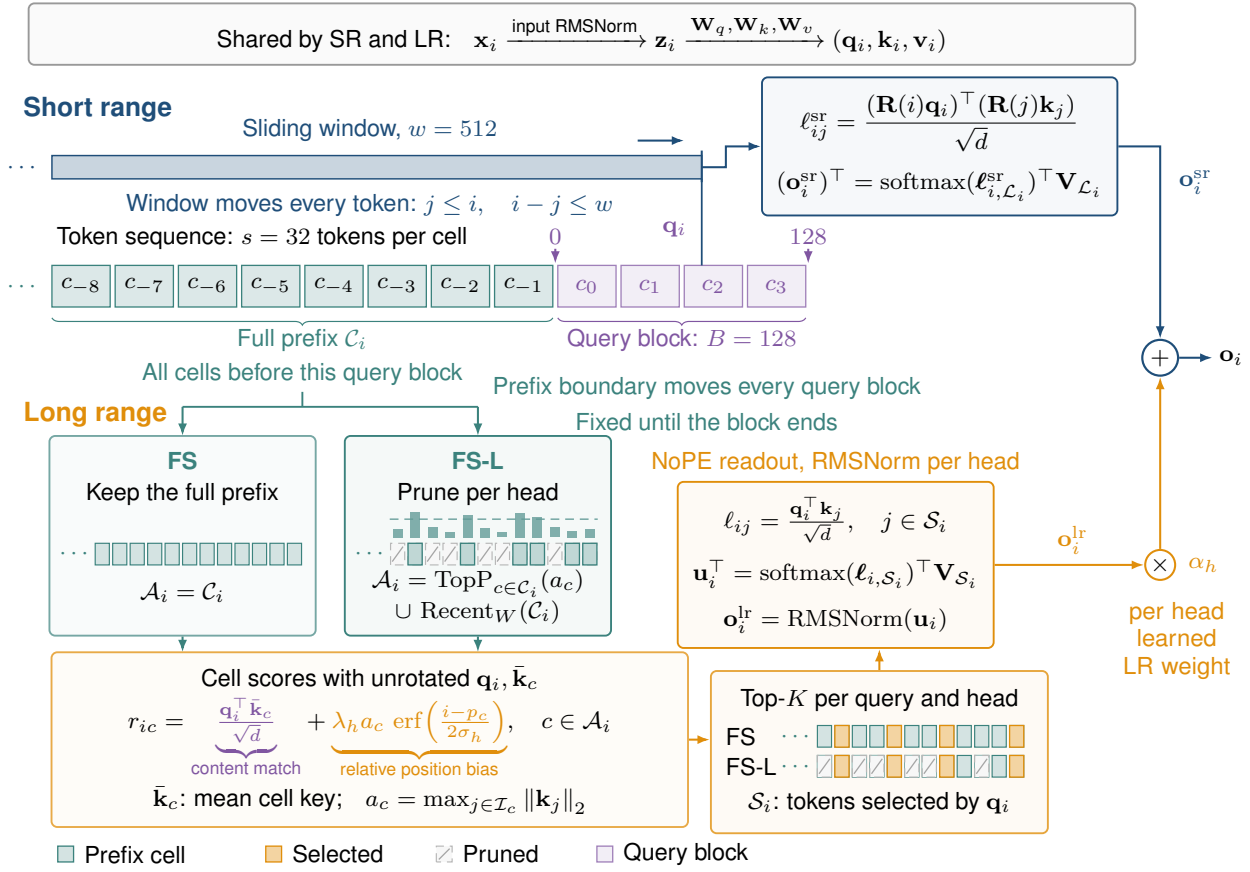

\centering
\definecolor{mypool}{HTML}{3B827D}
\definecolor{myquery}{HTML}{8056A1}
\resizebox{\linewidth}{!}{\begin{tikzpicture}[x=1cm,y=1cm,font=\sffamily\fontsize{8}{9.4}\selectfont,
  flow/.style={-{Latex[length=1.5mm,width=1.1mm]},draw=myblue,line width=.7pt},
  globalflow/.style={flow,draw=myorange},
  poolflow/.style={flow,draw=mypool},
  queryflow/.style={flow,draw=myquery},
  panel/.style={rounded corners=2pt,line width=.65pt},
  title/.style={font=\sffamily\bfseries\fontsize{9}{10}\selectfont,anchor=west},
  small/.style={font=\sffamily\fontsize{8}{9.4}\selectfont},
  every node/.style={inner sep=1.5pt}
]
\path[use as bounding box] (0,-.5) rectangle (15,10.10);

\node[panel,draw=black!40,fill=black!2,minimum width=13.30cm,
  minimum height=.65cm,inner xsep=5pt,inner ysep=4pt]
  at (6.85,9.63)
  {Shared by SR and LR:\quad
   $\vb x_i\xrightarrow{\text{input RMSNorm}}\vb z_i
    \xrightarrow{\vb W_q,\vb W_k,\vb W_v}(\vb q_i,\vb k_i,\vb v_i)$};

\node[title,text=myblue] at (.08,8.72) {Short range};
\node[small,text=myblue] at (4.30,8.48) {Sliding window, $w=512$};
\node[small,text=myblue] at (.14,8.03) {$\cdots$};
\draw[draw=myblue,fill=myblue!24,line width=.5pt]
  (.48,7.89) rectangle (8.29,8.17);
\draw[myblue,line width=.7pt] (8.29,7.85) -- (8.29,8.23);
\draw[flow] (7.52,8.35) -- (8.08,8.35);
\node[small,text=myblue] at (4.30,7.58)
  {Window moves every token: $j\le i,\quad i-j\le w$};
\node[panel,draw=myblue,fill=myblue!4,minimum width=4.00cm,
  minimum height=1.20cm,align=center,inner xsep=5pt,inner ysep=5pt]
  (srread) at (11.15,8.28)
  {$\ell_{ij}^{\mathrm{sr}}=
     \dfrac{(\vb R(i)\vb q_i)^\top(\vb R(j)\vb k_j)}{\sqrt d}$\\[5pt]
   $(\vb o_i^{\mathrm{sr}})^\top=
     \operatorname{softmax}(\boldsymbol\ell_{i,\mathcal L_i}^{\mathrm{sr}})^\top
     \vb V_{\mathcal L_i}$};
\draw[flow] (8.29,8.03) -- (8.58,8.03) |- (srread.west);

\input{figures/causal_availability}
\input{figures/long_range_routers}

\node[circle,draw=myblue,line width=.75pt,minimum size=.43cm,inner sep=0pt]
  (sum) at (13.80,5.74) {$+$};
\node[circle,draw=myorange,line width=.7pt,minimum size=.36cm,inner sep=0pt]
  (scale) at (13.80,3.25) {$\times$};
\draw[flow] (srread.east) -- (13.80,8.28) -- (sum.north);
\node[small,text=myblue,anchor=west] at (13.97,7.85) {$\vb o_i^{\mathrm{sr}}$};
\draw[globalflow] (lrread.east) -- (scale.west);
\node[small,text=myorange] at (12.75,3.55) {$\vb o_i^{\mathrm{lr}}$};
\node[small,anchor=west,text=myorange] at (14.08,3.25) {$\alpha_h$};
\node[small,anchor=north,align=center,text=myorange,text width=1.90cm]
  at (14.00,2.86) {per head learned\\LR weight};
\draw[globalflow] (scale.north) -- (sum.south);
\draw[flow] (sum.east) -- (14.43,5.74)
  node[right,inner sep=2pt] {$\vb o_i$};
\end{tikzpicture}}
\caption{\textbf{FreeSolo attention.}
FS scores the full prefix; FS-L prunes per head, independently of the query.
Both select $K=4$ cells per query and head. We use $P=64$ high-norm cells,
$W=8$ recent cells, cell size $s=32$, query blocks of $B=128$ tokens,
and SR radius $w=512$.}
\label{fig:arch}
\end{figure}

We introduce \textsc{FreeSolo} (FS), \emph{a native sparse mechanism that learns token
keys whose simple statistics make memory searchable, prunable, and hierarchically
summarizable}. Mean keys give a simple chunk representation for routing and
hierarchical search, while maximum token-key norms provide pruning priorities.
These summaries need only identify evidence: top-$K$
selection is followed by attention over the cells' original token keys and values.
FS scores all eligible cell summaries during training, but only selected token
values enter the long-range readout. The read budget stays fixed as history grows.
RoPE~\citep{su2021roformer}
stays within a fixed local window, while long-range routing and readout use
unrotated queries and keys, avoiding unseen rotary offsets (Figure~\ref{fig:arch}).

We develop two main variants through short CPT from base FS at the same
$4$k context length. FreeSolo-Linear (FS-L) uses norm pruning for $O(N)$
attention cost and bounded state. FreeSolo-Balanced Tree Search (FS-B)
keeps a growing archive and uses two content forests, with $O(N\log^2 N)$
algorithmic cost including radix-sort construction, superlinear but strictly
subquadratic (Section~\ref{sec:complexity}). Both extrapolate zero-shot to $16$k: at $800$M,
the RULER mean decreases only from $32.0$ to $22.3$ for FS-L and from
$37.7$ to $30.4$ for FS-B relative to $4$k. Notably, both retrieve single
needles at $256$k, $64\times$ the training length. At both scales, FS-B
achieves the highest RULER mean in the superlinear group at $4$k and beyond,
and FS-L leads the linear group at every reported length
(Table~\ref{tab:ruler}).

\section{Preliminaries}
\label{sec:prelim}
\paragraph{Attention.}
For one head, let $\vb{q}_i,\vb{k}_i,\vb{v}_i\in\R^d$ be the query, key,
and value at position $i\in\{0,\ldots,N-1\}$. Define
$\ell_{ij}=\vb{q}_i^\top\vb{k}_j/\sqrt d$ and $\Delta=i-j$.
For a nonempty token set $\mathcal J$, collect the logits in $\boldsymbol\ell_{i,\mathcal J}$
and matching value rows $\vb{v}_j^\top$ in $\vb{V}_{\mathcal J}$. The readout is
\begin{equation}
  \label{eq:std-attn}
  \vb{o}_i^\top
  =\softmax(\boldsymbol\ell_{i,\mathcal J})^\top\vb{V}_{\mathcal J}
  =\frac{\sum_{j\in\mathcal J}\exp(\ell_{ij})\vb{v}_j^\top}
         {\sum_{j\in\mathcal J}\exp(\ell_{ij})}.
\end{equation}
Full causal attention uses $\mathcal J=\{0,\ldots,i\}$ and costs $O(N^2d)$ per
sequence or $O(Nd)$ per cached decoding step. FlashAttention computes the same
readout with $O(Nd)$ memory~\citep{dao2022flash}.

\paragraph{Sparse attention.}
Sparse attention restricts each query to $\mathcal J_i\subseteq\{0,\ldots,i\}$,
fixed or query dependent. If $|\mathcal J_i|\le m$, the readout costs
$O(Nmd)$ per sequence, plus selection.

\paragraph{Length extrapolation.}
Zero-shot length extrapolation evaluates at $N>L$ after training at length $L$,
with fixed weights. We measure retrieval rather than perplexity, whose token
average can obscure errors requiring distant context~\citep{fang2024longppl}.

\methodpagebreak
\section{Method}
\label{sec:method}

\subsection{Range separation}
\label{sec:range-split}
FreeSolo combines local window attention with sparse long-range attention in each
head (Figure~\ref{fig:arch}). The local branch applies RoPE within a fixed window.
The long-range branch selects top-$K$ cells and applies NoPE
attention~\citep{kazemnejad2023nope} to their tokens, using unrotated queries and keys. This branch is sparse in both FS
and FS-L; FS-L also prunes the prefix before query-dependent scoring.
Both branches share projections of the layer's RMS-normalized hidden state
$\vb z_i$~\citep{zhang2019rmsnorm}: $\vb q_i=\vb W_q\vb z_i$,
$\vb k_i=\vb W_k\vb z_i$, and $\vb v_i=\vb W_v\vb z_i$. We combine the outputs as
\begin{equation}
  \label{eq:combine}
  \vb{o}_i=\vb{o}_i^{\mathrm{sr}}+\alpha_h\vb{o}_i^{\mathrm{lr}},
\end{equation}
where $\alpha_h$ is learned per head. Each branch normalizes its own attention,
and their supports may overlap. We concatenate the head outputs, project by
$\vb W_o$, and add the layer input. A second RMSNorm and
SwiGLU feed-forward residual~\citep{shazeer2020glu} complete the layer.

\subsection{Local branch}
\label{sec:local}
Let $\mathcal L_i=\{j:0\le j\le i,\ i-j\le w\}$, where $w$ is a look-back radius,
so a full window contains $w+1$ tokens. Let $\vb{R}(i)$ be the RoPE rotation at
position $i$~\citep{su2021roformer}. We use the local score
\begin{equation}
  \label{eq:local-score}
  \ell_{ij}^{\mathrm{sr}}
  =\frac{(\vb{R}(i)\vb{q}_i)^\top(\vb{R}(j)\vb{k}_j)}{\sqrt d}
  =\frac{\vb{q}_i^\top\vb{R}(j-i)\vb{k}_j}{\sqrt d},
\end{equation}
and collect these logits over $\mathcal L_i$ to compute
\begin{equation}
  \label{eq:local}
  (\vb{o}_i^{\mathrm{sr}})^\top
  =\softmax(\boldsymbol\ell_{i,\mathcal L_i}^{\mathrm{sr}})^\top\vb{V}_{\mathcal L_i}.
\end{equation}
We use $w=512$ at both training and evaluation. Increasing the context therefore
introduces no larger relative offsets into this branch.

\subsection{Long-range branch}
\label{sec:longrange}\label{sec:qqr}
The long-range branch provides a sparse, content-addressable memory beyond the
local window. Storing contextualized token key/value pairs allows later queries
to retrieve individual details directly~\citep{wu2022memorizing,xiao2024infllm}.
Mean keys guide cell selection; the readout uses the selected tokens' keys and values.
FS scores the full prefix before the query block. FS-L first prunes this prefix
by key norm and recency, separating which cells remain available from which
cells a query reads. This separates memory capacity from per-query read
bandwidth (Appendix~\ref{app:routing-probe}).

\paragraph{Cells.}
A cell is a group of $s$ consecutive token key/value pairs, selected as a unit by the
router. Using zero-based indices, cell $c$ contains positions
$\mathcal I_c=\{cs,\ldots,(c+1)s-1\}$. Its summaries are
\begin{equation}
  \label{eq:cells}
  \bar{\vb{k}}_c=\frac{1}{s}\sum_{j\in\mathcal I_c}\vb{k}_j,
  \qquad p_c=(c+\tfrac12)s,
  \qquad a_c=\max_{j\in\mathcal I_c}\norm{\vb{k}_j}_2.
\end{equation}
The mean key summarizes content for routing, $p_c$ lies halfway between the cell boundaries,
and the largest key norm supplies a query-independent priority. The raw keys and values remain
available for token-level attention after selection.

\paragraph{Query blocks.}
A query block groups $B/s$ consecutive cells, covering $B$ token positions, with $B$ a
multiple of $s$. The full prefix is shared across heads and queries in a block.
FS-L reuses each head's retained pool throughout the block, amortizing pruning
and gathering. Selection and readout are per query and head.
We use $B=128$ and $s=32$, so each block contains four cells. The block
containing query $i$ starts at $b_i=B\lfloor i/B\rfloor$.

\paragraph{Causal availability.}
The full prefix $\mathcal C_i$ contains all cells before the query-block start:
\begin{equation}
  \label{eq:available}
  \mathcal C_i=\{c\in\mathbb Z_{\ge0}:(c+1)s\le b_i\}.
\end{equation}
Every cell in this prefix contains only past tokens for every query in the block.
The prefix stays fixed until all $B$ queries have finished (Figure~\ref{fig:arch}).
At the next boundary, it includes the completed block, and FS-L updates its
retained cells. The local window covers the intervening past
because $w=512\ge B-1$; the first block has no long-range cells.

\paragraph{Cell scoring.}
We combine mean-key content routing, also used in MoBA~\citep{lu2025moba}, with
a norm-weighted relative position bias:
\begin{equation}
  \label{eq:qqr-score}
  r_{ic}=\frac{\vb{q}_i^\top\bar{\vb{k}}_c}{\sqrt d}
       +\lambda_h\,a_c\,g_h(i-p_c),
  \qquad c\in\mathcal C_i,
\end{equation}
where $g_h(\Delta)=\erf(\Delta/(2\sigma_h))$, $\sigma_h>0$, and $\lambda_h\ge0$.
The content term uses unrotated queries and keys. The erf prior encodes relative
distance in a translation-invariant way and supplies a norm-scaled bias to cell scoring.
All eligible offsets satisfy $i-p_c>0$. Ranking these scores and selecting only
the top-$K$ cells induces sparsity: only tokens in the selected cells enter the
long-range readout. FS scores the full prefix; FS-L first prunes it by key norm.

\paragraph{Query-independent pruning.}
Each head prunes independently using its own projected key norms.
FS-L retains the $W$ most recent cells and the $P$ cells with largest $a_c$:
\begin{equation}
  \label{eq:pool}
  \mathcal A_i=\operatorname{Recent}_W(\mathcal C_i)
       \ \cup\ \operatorname{TopP}_{c\in\mathcal C_i}(a_c),
  \qquad |\mathcal A_i|\le W+P.
\end{equation}
The union removes duplicates. The priority favors cells containing a large-norm
key, while the recent window preserves nearby context. Projected keys have
varying norms; we do not normalize them individually. The RMSNorm after the
long-range readout normalizes the output, not these priorities.

Key norm supplies a query-independent priority, not a guarantee of relevance
to every query. CPT adapts the model to this restricted access pattern.
Appendix~\ref{app:routing-probe} measures how much pruning compresses the prefix
and which cells remain available for retrieval. We use $P=64$ and $W=8$.

\paragraph{Per-query selection.}
FS uses $\mathcal A_i=\mathcal C_i$; FS-L uses Eq.~\eqref{eq:pool}.
Each head evaluates Eq.~\eqref{eq:qqr-score} on its own $\mathcal A_i$ and selects
up to $K=4$ cells per query. Different heads can select different cells, and
selections can change at every token within a query block.
Let $\mathcal T_i(\mathcal A_i)$ contain
the $\min(K,|\mathcal A_i|)$ highest-scoring cells, and collect their token indices as
$\mathcal S_i=\bigcup_{c\in\mathcal T_i(\mathcal A_i)}\mathcal I_c$.

\paragraph{Token readout.}
The long-range output is
\begin{equation}
  \label{eq:glob}
  \vb{u}_i^\top=\softmax(\boldsymbol\ell_{i,\mathcal S_i})^\top\vb{V}_{\mathcal S_i},
  \qquad \vb{o}_i^{\mathrm{lr}}=\operatorname{RMSNorm}(\vb{u}_i).
\end{equation}
This is NoPE attention: the erf prior affects cell scoring only, while the readout
uses unmodified token keys and values, with no positional bias in its logits.
The softmax is normalized jointly over all selected tokens.
RMSNorm acts over the $d$ head features~\citep{zhang2019rmsnorm}. Both outputs in
Eq.~\eqref{eq:glob} are zero when $\mathcal S_i$ is empty.

\paragraph{Training.}
We obtain FS-L by continued pretraining of FS with pruning enabled.
Both stages use the standard next-token cross-entropy objective on
$4096$-token sequences. CPT resets the optimizer and learning-rate schedule
(Appendix~\ref{app:hyperparams}) and requires no longer training contexts.
Causal masks allow all query blocks to be evaluated in parallel. Hard top-$P$
and top-$K$ indices receive no gradient; shared projections and contextual
states learn through local attention and selected-token readout. The
routing-only scalars receive no gradient in this configuration and remain
at $\sigma_h=4$ and $\lambda_h=1$.

Training with pruning from scratch is also possible, but CPT gives better
overall retrieval, particularly on the harder tasks. The $340$M from-scratch
model retains strong S-NIAH-1 recall, whereas the tested $800$M run retrieves
poorly. Starting from FS provides useful representations before imposing the
pruning constraint. This motivates the two-stage recipe, although the runs do
not isolate the cause of the from-scratch difference.
Appendix~\ref{app:sparse-fail} gives the comparisons and variation across CPT runs.

\paragraph{Inference.}
Prompt prefill uses the same causal operator and populates each layer's raw
key/value cache and cell summaries. During decoding, each new token appends its
key and value; a completed cell supplies its mean key and maximum key norm.
The local window advances every token. The long-range prefix advances only at
query-block boundaries, when FS-L also updates its retained set. Within a block,
each query scores the available cells, selects its own top-$K$, and applies
Eq.~\eqref{eq:glob} before the branch combination in Eq.~\eqref{eq:combine}.
FS-L's streaming cache merges newly eligible cells into the retained set and
discards unused KV once it leaves the recent, local, and unfinished-block buffers.

\subsection{Balanced tree search (FS-B)}
\label{sec:fsb}
\begin{figure}[t]
\centering
\input{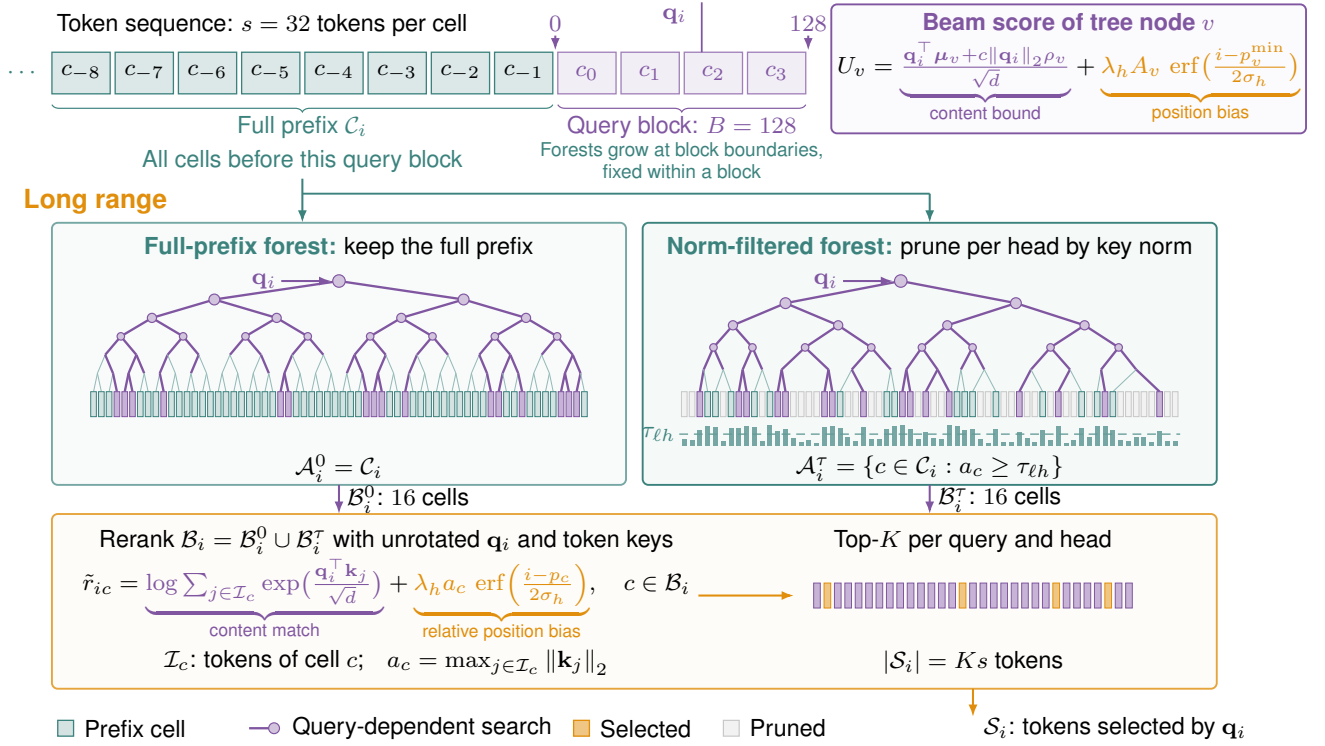}
\caption{\textbf{FS-B long-range selection, per head.}
The query searches two causal forests over the prefix: one over every cell
and one over cells with $a_c\ge\tau_{\ell h}$. Each beam keeps $16$ nodes
per level down to the leaves, $\lceil\log_2 M\rceil$ levels for $M$
admitted cells ($6$ and $5$ in this $64$-cell example). The union of both
beams is reranked with token keys; the top $K=4$ cells form $\mathcal S_i$,
which is read as in FS.}
\label{fig:fsb-selection}
\end{figure}
FS-L suits evidence concentrated in a few cells, as in single-needle retrieval,
because query-independent pruning can keep those cells. Dense, independently
queryable facts are harder: in a list of unique key--value pairs followed by a
question about a uniformly sampled key, every pair is equally likely to matter,
so a fixed pool reaches a shrinking fraction of the possible targets
(Appendix~\ref{app:learned-retention}). FS keeps every cell reachable but scores
the full prefix in $O(N^2/s)$ work. FS-B keeps the local branch and the NoPE
readout of Eq.~\eqref{eq:glob} and changes only how $\mathcal S_i$ is chosen:
each query searches content trees over a growing memory, so no cell is
discarded for all future queries while search and read budgets stay bounded.

\paragraph{Two causal forests.}
FS-B always uses two forests per head, with admitted sets
\begin{equation}
  \label{eq:fsb-admission}
  \mathcal A_i^0=\mathcal C_i,\qquad
  \mathcal A_i^\tau=\{c\in\mathcal C_i:a_c\ge\tau_{\ell h}\},
\end{equation}
where $\tau_{\ell h}$ is the median cell norm on FineWeb-Edu windows,
calibrated once per checkpoint and reused at all lengths. The first forest
keeps the whole prefix searchable; the second emphasizes high-norm evidence,
as FS-L does, without a fixed quota. In each forest, every aligned chunk of $2^k$
admitted cells is split recursively at the median of its mean keys along
their principal direction, and summary nodes join the chunk roots, so $M$
admitted cells give height $\lceil\log_2 M\rceil$
(Figure~\ref{fig:fsb-construction}). A node $v$ over cells $\mathcal C(v)$
stores their mean key $\boldsymbol\mu_v$, a radius
$\rho_v\ge\max_{c\in\mathcal C(v)}\norm{\bar{\vb k}_c-\boldsymbol\mu_v}_2$,
the maximum norm $A_v=\max_{c\in\mathcal C(v)}a_c$, and the earliest
position $p_v^{\min}=\min_{c\in\mathcal C(v)}p_c$.

\paragraph{Beam search.}
From the root, each forest keeps the $W^{\mathrm b}=16$ highest-priority
nodes per level down to the leaves, ranking nodes by
\begin{equation}
  \label{eq:fsb-bound}
  U_v=\frac{\vb q_i^\top\boldsymbol\mu_v+c\norm{\vb q_i}_2\rho_v}{\sqrt d}
      +\lambda_h A_v\,g_h(i-p_v^{\min}).
\end{equation}
For $c\ge1$, $U_v$ bounds the score of Eq.~\eqref{eq:qqr-score} for every
cell below $v$: the radius bounds the content term, and $A_v$ and
$p_v^{\min}$ bound the nonnegative, monotone prior. The search remains
approximate, since a branch with a lower bound can still hold the best
cell; we use $c=8$. The leaves reached form candidate sets
$\mathcal B_i^0$ and $\mathcal B_i^\tau$ of at most $16$ cells each, which
differ because the forests group the prefix differently.

\paragraph{Token-level reranking.}
Mean keys can hide a single useful token, so FS-B rescores the deduplicated
candidates $\mathcal B_i=\mathcal B_i^0\cup\mathcal B_i^\tau$ with their token keys,
\begin{equation}
  \label{eq:fsb-rerank}
  \tilde r_{ic}=\log\sum_{j\in\mathcal I_c}
     \exp\!\Big(\frac{\vb q_i^\top\vb k_j}{\sqrt d}\Big)
     +\lambda_h a_c\,g_h(i-p_c),
  \qquad c\in\mathcal B_i,
\end{equation}
and the top $K=4$ cells form $\mathcal S_i$, read with Eq.~\eqref{eq:glob}.

\paragraph{Retrieval-weighted CPT.}
At $340$M, FS-B reuses the FS-L weights. At $800$M, base FS undergoes
$0.42$B-token CPT on $4096$-token FineWeb-Edu sequences with a weighted
next-token loss:
\begin{equation}
  \label{eq:fsb-recall-loss}
  \mathcal L=-\frac{\sum_t w_t\log p_\theta(x_{t+1}\mid x_{\le t})}{\sum_t w_t},
  \qquad w_t=1+\beta f_t+\beta_{\mathrm{sw}}\chi_t,
\end{equation}
Here $f_t,\chi_t\in\{0,1\}$ are text-derived masks. The recall mask $f_t=1$
marks starts and continuations of repeated spans, identified by a unique
earlier four-token match more than $512$ tokens back in the same document.
The switch mask $\chi_t=1$ adds a bonus at a recall start within $64$ tokens
of the preceding start if their earlier source positions are more than
$64$ tokens apart. Every token retains base weight $1$. Training uses full-prefix
FS selection with $16$-cell reads; forests are used only at inference
(Appendix~\ref{app:fsb-cpt}).

\subsection{Computational cost}
\label{sec:complexity}
At fixed head dimensions and budgets, FS costs $O(N^2/s)$, FS-L $O(N)$,
and FS-B $O(N\log^2 N)$ including radix-sort construction.

Cell statistics cost $O(Nd)$ per head. Each query scores $M\simeq N/s$ cells,
giving FS a cost of $O(N^2d/s+N(w+Ks)d)$ including local and selected-token
readout. Query blocks share availability, not scores; $B$ does not reduce
the quadratic term. FS-L incrementally merges newly eligible cells into its
top-$P$ set. Fixed key-norm scores prevent discarded cells from re-entering.
A heap costs $O(M\log(P+1))$ in total; bounded-array merges are linear at fixed budgets.

Let $C=P+W$. Scoring the retained cells costs $O(NCd)$, and attention over gathered
selected tokens costs $O(NKsd)$. With incremental retention and the local branch, the sequence cost is
\begin{equation}
  \label{eq:cost}
  O\!\left(Nd+M\log(P+1)+N(C+w+Ks)d\right).
\end{equation}
At fixed budgets, decode work is bounded, including amortized retention updates.
The streaming implementation keeps $O(((P+W)s+w+B+s)d)$ token state per head
plus indices and counters, computes rotations on demand, and prefills in fixed
chunks (Appendix~\ref{app:bounded-cache}).

FS-L results in Tables~\ref{tab:ruler} and~\ref{tab:ruler-full} use this bounded cache, which keeps
only the retained cells (Appendix~\ref{app:retrieval-setup}); its resource
measurements are in Appendix~\ref{app:bounded-cache}. The efficiency
comparison in Table~\ref{tab:eff-infer} uses a masked implementation that
stores $O(Nd)$ KV and rescans priorities at block boundaries, adding
$\Theta(N^2/(Bs))$ work, with $O(N(P+W)sd)$ masked readout.

FS-B searches $M\le\lfloor N/s\rfloor=\lfloor N/32\rfloor$ admitted cells
per forest. Beams descend $\lceil\log_2 M\rceil$ levels; search, reranking
and readout cost $O(d[W^{\mathrm b}\log M+W^{\mathrm b}s+Ks])$ per query,
or $O(N\log N)$ per sequence. Fixed-width radix sorting adds
$O(Md\log^2 M)$ construction work, giving $O(N\log^2 N)$ combined
algorithmic cost and $O(Nd)$ archive space per head.
Appendix~\ref{app:fsb-cost} derives this bound and discloses repeated
buffer copies in the reference streaming cache.

\section{Experiments}
\label{sec:exp}

\subsection{Setup}
\label{sec:setup}
We train LLaMA-style decoders~\citep{touvron2023llama} on
FineWeb-Edu~\citep{penedo2024fineweb}, using a $32$k LLaMA tokenizer and a
$4096$-token context. The two FreeSolo configurations have $341$M and $743$M
parameters, with pretraining budgets of $15$B and $32$B tokens. Training uses bf16 and AdamW~\citep{loshchilov2019adamw};
see Appendix~\ref{app:hyperparams} for architecture and optimizer settings.

FS-L uses $0.73$B/$1.57$B CPT tokens at $340$M/$800$M, about $5\%$ of
pretraining; FS-B uses $0.42$B at $800$M (Section~\ref{sec:fsb}). All CPT uses
$4096$-token contexts. Main-table baselines use pretraining checkpoints.
Appendix~\ref{app:sparse-fail} adds selected $800$M-scale baseline CPT
controls, FS/FS-L comparisons at both scales, and seed and from-scratch studies.

\paragraph{Baselines.}
TF++ is the full-attention Transformer++ reference~\citep{yang2024gatedeltanet}
with RoPE; SWA uses a $512$-token RoPE window. NSA~\citep{yuan2025nsa} and
MoBA~\citep{lu2025moba} select blocks for sparse token attention.
GDN~\citep{yang2024gatedeltanet} uses a recurrent gated delta rule, and
NAtS-L~\citep{deng2026natsl} combines GDN with retained-token softmax attention.
InfLLM~\citep{xiao2024infllm} adds inference-time retrieval to the TF++ checkpoint;
we evaluate it on S-NIAH-1/2/3.

Raven~\citep{afzal2026raven} writes sparsely to key/value memory slots and reads
them with softmax. We use the standalone $256$-slot model as the linear-time
Raven baseline. Raven-full has stronger published S-NIAH-2/3 recall, but its
full NoPE attention costs $O(N^2)$, so we exclude it from this linear-time comparison. Its SWA hybrid is
linear at fixed window size but has weaker recall than Raven-full~\citep[Table~9]{afzal2026raven}.
Baselines share the corpus, tokenizer, and training context within each scale.
The tables list parameter counts; Appendix~\ref{app:hyperparams} records training budgets and settings.

\paragraph{Complexity.}
Table~\ref{tab:ruler} reports algorithmic attention complexity at fixed head dimensions
and budgets: $O(N^2/s)$ for FS, $O(N^2/b)$ for MoBA and NSA with a fixed
block factor $b$, $O(N\log^2 N)$ for FS-B including radix-sort construction, and $O(N)$ for FS-L
(Section~\ref{sec:complexity}), Raven, GDN, and SWA. NAtS-L costs $O(NR)$ for $R$ retained tokens, quadratic at a
fixed retention fraction~\citep{deng2026natsl}. Our InfLLM lookup is exhaustive.

\paragraph{Evaluation.}
We evaluate NVIDIA RULER~\citep{hsieh2024ruler} with $100$ examples per task and
length, greedy decoding, and official $0$--$100$ scores. Suite means cover
$12$ tasks at $2$k/$4$k and $13$ at $8$k/$16$k, which add QA-2.
We omit further long-context runs when a model's retrieval has already failed
at shorter extrapolation lengths.
Appendix~\ref{app:ruler-full} reports all task scores; Appendix~\ref{app:hyperparams}
gives the protocol.

\subsection{Zero-shot extrapolation for long-context retrieval}
\label{sec:extrap}\label{sec:full-suite}
Raven, NAtS-L, and GDN retain retrieval ability beyond the training context,
whereas MoBA and NSA do not sustain recall in our evaluations (Table~\ref{tab:ruler}).
FreeSolo's advantage is clearest on the harder S-NIAH-2/3 tasks
(Figure~\ref{fig:extrap}). At $800$M and $16$k, four times the training length,
FS-L scores $81/40$, compared with $26/0$ for Raven and $72/34$ for FS.
FS-L reaches $53/13$ at $64$k, compared with $28/4$ for FS and $0/0$ for
NAtS-L. Relative to FS, the $800$M FS-L checkpoint trades some S-NIAH-1 recall
for higher S-NIAH-2/3 scores at $16$k--$64$k, although absolute accuracy
still declines with length.

FS-B's two-forest search (Section~\ref{sec:fsb}) improves retrieval further:
at $800$M and $16$k it scores $100/88/84$ on S-NIAH-1/2/3, with a full RULER
mean of $30.4$.

FS and FS-L also outperform Raven and NAtS-L on the full RULER mean at both scales.
At $800$M/$16$k, the means are $23.1/22.3$ for FS/FS-L, versus $12.8/2.8$
for Raven/NAtS-L. The gains extend beyond single needles: FS-L scores $23$
on MK-NIAH-1 versus $3$ for both linear baselines, and $17$ on MV-NIAH versus
$9/0.5$ for Raven/NAtS-L (Table~\ref{tab:ruler-full}). Some tasks, including
variable tracking, remain difficult for all models even within the training context.

Matched continuations distinguish adaptation to pruning from additional
training (Appendix~\ref{app:sparse-fail}). On the same $50$ prompts at
$800$M/$16$k, FS continuation scores $84/38$ on S-NIAH-2/3 and FS-L scores
$80/42$. Thus short CPT preserves much of the harder-task retrieval while
reducing attention cost from quadratic to linear.

S-NIAH-1 tests retrieval over a much longer range. At $340$M, FS maintains
$98$--$100$ through $256$k and FS-L scores $84$ at $256$k; Raven falls to
$20$ at $131$k. At $800$M, FS and FS-L score $96/85$ at $256$k, compared
with zero for NAtS-L. Raven also extrapolates well on this task, scoring
$99$--$100$ through $131$k and $92$ at $256$k.

\begin{table}[t]
\caption{\textbf{RULER accuracy} ($0$--$100$).
All models train at $4$k; results at $8$k and above test zero-shot extrapolation.
Means cover available tasks; cost is defined in Section~\ref{sec:setup}.
FS-B's algorithmic cost includes radix-sort construction (Appendix~\ref{app:fsb-cost}).
Bold/underline mark the best/second-best scores within each block (super-linear or linear cost);
dashes denote unavailable results.}
\label{tab:ruler}\label{tab:extrap}\label{tab:ruler-suite}
\centering
\mainresultstablefont\setlength{\tabcolsep}{0.3pt}
\begin{tabular*}{\textwidth}{@{\extracolsep{\fill}}lrc@{\hspace{1pt}}cccccccc@{\hspace{1pt}}cccccc@{\hspace{1pt}}cccccc@{\hspace{1pt}}cccc@{}}
\toprule
& & & \multicolumn{8}{c}{S-NIAH-1} & \multicolumn{6}{c}{S-NIAH-2} & \multicolumn{6}{c}{S-NIAH-3} & \multicolumn{4}{c}{Mean} \\
\cmidrule(lr){4-11}\cmidrule(lr){12-17}\cmidrule(lr){18-23}\cmidrule(lr){24-27}
Model & Params & Complexity & 2k & 4k & 8k & 16k & 32k & 64k & 131k & 256k & 2k & 4k & 8k & 16k & 32k & 64k & 2k & 4k & 8k & 16k & 32k & 64k & 2k & 4k & 8k & 16k \\
\midrule
TF++ & $335$M & $O(N^2)$ & $95$ & $57$ & $0$ & $0$ & -- & -- & -- & -- & $\underline{91}$ & $24$ & $0$ & $0$ & -- & -- & $38$ & $1$ & $0$ & $0$ & -- & -- & $26.2$ & $13.3$ & $0.2$ & $0.0$ \\
InfLLM & $335$M & $O(N^2)$ & $94$ & $\underline{58}$ & $13$ & $9$ & -- & -- & -- & -- & $89$ & $23$ & $8$ & $2$ & -- & -- & $39$ & $1$ & $2$ & $\underline{1}$ & -- & -- & -- & -- & -- & -- \\
MoBA & $341$M & $O(N^2/b)$ & $76$ & $12$ & $0$ & $0$ & -- & -- & -- & -- & $83$ & $23$ & $0$ & $0$ & -- & -- & $\underline{59}$ & $3$ & $0$ & $0$ & -- & -- & $25.6$ & $7.6$ & $0.0$ & $0.0$ \\
NSA & $513$M & $O(N^2/b)$ & $43$ & $23$ & $0$ & $0$ & -- & -- & -- & -- & $48$ & $18$ & $0$ & $0$ & -- & -- & $1$ & $0$ & $0$ & $0$ & -- & -- & $13.2$ & $7.3$ & $0.9$ & $0.0$ \\
Raven-full & $355$M & $O(N^2)$ & $\mathbf{100}$ & $\mathbf{99}$ & $83$ & $39$ & $44$ & $\underline{4}$ & $0$ & $\underline{0}$ & $\mathbf{100}$ & $\underline{89}$ & $\underline{41}$ & $7$ & -- & -- & $38$ & $1$ & $0$ & $0$ & -- & -- & $26.7$ & $21.5$ & $13.5$ & $7.1$ \\
\textbf{FS} & $341$M & $O(N^2/s)$ & $\underline{99}$ & $\mathbf{99}$ & $\mathbf{100}$ & $\mathbf{100}$ & $\underline{98}$ & $\mathbf{100}$ & $\underline{98}$ & $\mathbf{100}$ & $\mathbf{100}$ & $77$ & $\underline{41}$ & $\underline{17}$ & -- & -- & $58$ & $\underline{14}$ & $\underline{7}$ & $0$ & -- & -- & $\underline{30.8}$ & $\underline{23.2}$ & $\underline{16.5}$ & $\underline{11.8}$ \\
\textbf{FS-B} & $341$M & $O(N\log^2 N)$ & $\mathbf{100}$ & $\mathbf{99}$ & $\underline{99}$ & $\underline{99}$ & $\mathbf{99}$ & $\mathbf{100}$ & $\mathbf{100}$ & $\mathbf{100}$ & $\mathbf{100}$ & $\mathbf{99}$ & $\mathbf{89}$ & $\mathbf{60}$ & $\mathbf{32}$ & $\mathbf{10}$ & $\mathbf{91}$ & $\mathbf{78}$ & $\mathbf{64}$ & $\mathbf{27}$ & $\mathbf{34}$ & $\mathbf{24}$ & $\mathbf{34.2}$ & $\mathbf{31.1}$ & $\mathbf{26.6}$ & $\mathbf{20.2}$ \\
\midrule
\textbf{FS-L} & $341$M & $O(N)$ & $\underline{99}$ & $88$ & $\underline{90}$ & $\mathbf{87}$ & $\underline{65}$ & $\mathbf{86}$ & $\mathbf{85}$ & $\mathbf{84}$ & $\mathbf{99}$ & $\mathbf{70}$ & $\underline{34}$ & $\mathbf{20}$ & $\mathbf{7}$ & $\mathbf{3}$ & $\mathbf{70}$ & $\mathbf{17}$ & $\mathbf{9}$ & $\mathbf{1}$ & $\mathbf{2}$ & $\mathbf{1}$ & $\mathbf{32.4}$ & $\mathbf{21.6}$ & $\mathbf{15.5}$ & $\mathbf{10.7}$ \\
Raven & $357$M & $O(N)$ & $95$ & $\mathbf{96}$ & $\mathbf{93}$ & $\underline{81}$ & $\mathbf{78}$ & $\underline{41}$ & $\underline{20}$ & -- & $\underline{92}$ & $\mathbf{70}$ & $\mathbf{36}$ & $\underline{3}$ & -- & -- & $3$ & $0$ & $0$ & $\underline{0}$ & -- & -- & $20.5$ & $\underline{17.8}$ & $\underline{14.0}$ & $\underline{8.5}$ \\
NAtS-L & $380$M & $O(NR)$ & $\mathbf{100}$ & $\underline{94}$ & $37$ & $19$ & $0$ & $0$ & $0$ & -- & $89$ & $\underline{22}$ & $20$ & $\underline{3}$ & -- & -- & $\underline{36}$ & $\underline{6}$ & $1$ & $\underline{0}$ & -- & -- & $\underline{25.6}$ & $15.3$ & $7.4$ & $4.0$ \\
GDN & $335$M & $O(N)$ & $95$ & $83$ & $35$ & $19$ & -- & -- & -- & -- & $83$ & $20$ & $18$ & $\underline{3}$ & -- & -- & $0$ & $0$ & $0$ & $\underline{0}$ & -- & -- & $21.6$ & $14.2$ & $7.0$ & $3.6$ \\
SWA & $341$M & $O(N)$ & $18$ & $15$ & $7$ & $3$ & -- & -- & -- & -- & $31$ & $21$ & $6$ & $2$ & -- & -- & $25$ & $2$ & $\underline{3}$ & $\underline{0}$ & -- & -- & $14.6$ & $9.3$ & $4.5$ & $4.0$ \\
\midrule
\multicolumn{27}{@{}l}{\emph{$\approx\!800$M scale}} \\
\midrule
TF++ & $743$M & $O(N^2)$ & $\mathbf{100}$ & $95$ & $0$ & $0$ & -- & -- & -- & -- & $\mathbf{100}$ & $94$ & $0$ & $0$ & -- & -- & $67$ & $50$ & $0$ & $0$ & -- & -- & $\mathbf{46.3}$ & $\underline{37.5}$ & $0.0$ & $0.0$ \\
InfLLM & $743$M & $O(N^2)$ & $\mathbf{100}$ & $96$ & $20$ & $4$ & -- & -- & -- & -- & $\mathbf{100}$ & $89$ & $12$ & $4$ & -- & -- & $77$ & $50$ & $4$ & $0$ & -- & -- & -- & -- & -- & -- \\
NSA & $1128$M & $O(N^2/b)$ & $\mathbf{100}$ & $75$ & $8$ & $0$ & -- & -- & -- & -- & $\underline{98}$ & $46$ & $3$ & $0$ & -- & -- & $30$ & $7$ & $0$ & $0$ & -- & -- & $40.7$ & $24.8$ & $2.6$ & $0.1$ \\
Raven-full & $723$M & $O(N^2)$ & $\mathbf{100}$ & $\mathbf{100}$ & $\mathbf{100}$ & $\mathbf{100}$ & $\underline{98}$ & $91$ & $74$ & $48$ & $\mathbf{100}$ & $88$ & $81$ & $69$ & $51$ & $1$ & $\underline{91}$ & $58$ & $21$ & $4$ & -- & -- & $41.4$ & $33.7$ & $26.5$ & $\underline{23.8}$ \\
\textbf{FS} & $743$M & $O(N^2/s)$ & $\underline{97}$ & $91$ & $93$ & $\underline{95}$ & $79$ & $\underline{97}$ & $\underline{95}$ & $\underline{96}$ & $\mathbf{100}$ & $\underline{99}$ & $\underline{92}$ & $\underline{72}$ & $\underline{63}$ & $\underline{28}$ & $85$ & $\underline{71}$ & $\underline{51}$ & $\underline{34}$ & $\underline{27}$ & $\underline{4}$ & $39.4$ & $34.0$ & $\underline{27.6}$ & $23.1$ \\
\textbf{FS-B} & $743$M & $O(N\log^2 N)$ & $\underline{97}$ & $\underline{98}$ & $\underline{97}$ & $\mathbf{100}$ & $\mathbf{100}$ & $\mathbf{100}$ & $\mathbf{100}$ & $\mathbf{100}$ & $\mathbf{100}$ & $\mathbf{100}$ & $\mathbf{95}$ & $\mathbf{88}$ & $\mathbf{80}$ & $\mathbf{43}$ & $\mathbf{96}$ & $\mathbf{93}$ & $\mathbf{87}$ & $\mathbf{84}$ & $\mathbf{71}$ & $\mathbf{69}$ & $\underline{42.1}$ & $\mathbf{37.7}$ & $\mathbf{31.8}$ & $\mathbf{30.4}$ \\
\midrule
\textbf{FS-L} & $743$M & $O(N)$ & $\mathbf{100}$ & $\underline{92}$ & $\underline{84}$ & $\underline{84}$ & $\underline{84}$ & $\underline{78}$ & $\underline{80}$ & $\underline{85}$ & $\underline{99}$ & $\mathbf{99}$ & $\mathbf{89}$ & $\mathbf{81}$ & $\mathbf{68}$ & $\mathbf{53}$ & $\mathbf{90}$ & $\mathbf{71}$ & $\mathbf{55}$ & $\mathbf{40}$ & $\mathbf{29}$ & $\mathbf{13}$ & $\mathbf{37.2}$ & $\mathbf{32.0}$ & $\mathbf{25.6}$ & $\mathbf{22.3}$ \\
Raven & $726$M & $O(N)$ & $\mathbf{100}$ & $\mathbf{100}$ & $\mathbf{100}$ & $\mathbf{100}$ & $\mathbf{100}$ & $\mathbf{99}$ & $\mathbf{99}$ & $\mathbf{92}$ & $\underline{99}$ & $\underline{97}$ & $\underline{81}$ & $\underline{26}$ & $\underline{5}$ & -- & $32$ & $7$ & $1$ & $0$ & $\underline{0}$ & -- & $30.1$ & $\underline{23.9}$ & $\underline{18.4}$ & $\underline{12.8}$ \\
NAtS-L & $793$M & $O(NR)$ & $\underline{86}$ & $72$ & $59$ & $21$ & $2$ & $1$ & $0$ & $0$ & $\mathbf{100}$ & $76$ & $28$ & $3$ & $0$ & $\underline{0}$ & $62$ & $24$ & $6$ & $0$ & $\underline{0}$ & $\underline{0}$ & $27.8$ & $19.4$ & $10.2$ & $2.8$ \\
GDN & $694$M & $O(N)$ & $\mathbf{100}$ & $\mathbf{100}$ & $\mathbf{100}$ & $56$ & $22$ & -- & -- & -- & $94$ & $38$ & $19$ & $13$ & -- & -- & $\underline{84}$ & $\underline{27}$ & $\underline{10}$ & $\underline{5}$ & -- & -- & $\underline{35.8}$ & $22.8$ & $16.5$ & $10.4$ \\
SWA & $743$M & $O(N)$ & $18$ & $15$ & $7$ & $3$ & -- & -- & -- & -- & $31$ & $21$ & $6$ & $2$ & -- & -- & $19$ & $3$ & $1$ & $0$ & -- & -- & $15.6$ & $10.3$ & $5.5$ & $4.0$ \\
\bottomrule
\end{tabular*}
\end{table}

\subsection{Short-context quality}
\label{sec:downstream-zs}
Table~\ref{tab:downstream-zs} reports eight zero-shot tasks with task definitions
from the Language Model Evaluation Harness~\citep{gao2024harness}. We use \texttt{acc\_norm} for
HellaSwag, ARC-c, and OBQA, and \texttt{acc} for the other tasks; Avg is their
arithmetic mean. At $340$M, FS averages $51.9$ and FS-L $51.0$, compared with
$51.8$ for TF++, $51.4$ for Raven, and $50.8$ for NAtS-L. At $800$M, FS
reaches $56.5$, close to TF++'s $56.8$ and SWA's $56.2$, while FS-L reaches
$55.2$ and Raven $54.6$.
Both FS-L checkpoints retain similar short-context quality.

\begin{table}[t]
\caption{\textbf{Zero-shot downstream accuracy} ($0$--$100$).
Rows are ordered by average within each scale; bold and underline mark the best and second-best scores.}
\label{tab:downstream-zs}
\centering
\small\setlength{\tabcolsep}{4pt}
\begin{tabular}{@{}lrccccccccc@{}}
\toprule
Model & Params & HellaS. & PIQA & ARC-e & ARC-c & WinoG. & OBQA & SciQ & BoolQ & Avg \\
\midrule
GDN & $335$M & $\underline{39.2}$ & $\underline{66.8}$ & $55.8$ & $\mathbf{28.2}$ & $\underline{51.9}$ & $\underline{33.0}$ & $\mathbf{83.2}$ & $59.8$ & $\mathbf{52.2}$ \\
\textbf{FS} & $341$M & $38.9$ & $\underline{66.8}$ & $\mathbf{57.6}$ & $25.9$ & $50.9$ & $32.4$ & $81.2$ & $\mathbf{61.1}$ & $\underline{51.9}$ \\
TF++ & $335$M & $\underline{39.2}$ & $66.0$ & $56.2$ & $\underline{28.1}$ & $50.6$ & $32.8$ & $\underline{81.5}$ & $59.6$ & $51.8$ \\
Raven & $357$M & $37.6$ & $65.2$ & $55.8$ & $27.6$ & $51.2$ & $32.2$ & $80.6$ & $\underline{60.8}$ & $51.4$ \\
SWA & $341$M & $38.8$ & $66.2$ & $\underline{56.4}$ & $26.7$ & $50.6$ & $30.8$ & $80.4$ & $\mathbf{61.1}$ & $51.4$ \\
\textbf{FS-L} & $341$M & $38.4$ & $66.3$ & $55.0$ & $25.3$ & $51.0$ & $31.4$ & $79.8$ & $60.4$ & $51.0$ \\
\textbf{FS-B} & $341$M & $38.4$ & $66.3$ & $55.0$ & $25.3$ & $51.0$ & $31.4$ & $79.8$ & $60.3$ & $51.0$ \\
NAtS-L & $380$M & $\mathbf{39.9}$ & $\mathbf{67.6}$ & $50.0$ & $27.4$ & $\mathbf{52.3}$ & $\mathbf{34.0}$ & $75.1$ & $59.7$ & $50.8$ \\
MoBA & $341$M & $33.3$ & $61.8$ & $43.4$ & $26.2$ & $51.3$ & $29.0$ & $65.1$ & $52.2$ & $45.3$ \\
NSA & $513$M & $32.3$ & $62.6$ & $43.8$ & $24.9$ & $\underline{51.9}$ & $28.8$ & $66.4$ & $45.0$ & $44.5$ \\
\midrule
\multicolumn{11}{@{}l}{\emph{$\approx\!800$M scale}} \\
\midrule
TF++ & $743$M & $48.9$ & $\mathbf{70.9}$ & $63.9$ & $\underline{32.9}$ & $54.3$ & $34.4$ & $87.4$ & $\mathbf{61.9}$ & $\mathbf{56.8}$ \\
\textbf{FS} & $743$M & $\underline{49.5}$ & $\underline{70.0}$ & $\underline{64.6}$ & $31.7$ & $\mathbf{56.0}$ & $34.4$ & $86.7$ & $58.8$ & $\underline{56.5}$ \\
Raven-full & $723$M & $46.9$ & $69.4$ & $64.5$ & $\mathbf{33.6}$ & $54.1$ & $36.8$ & $85.9$ & $60.4$ & $\underline{56.5}$ \\
GDN & $694$M & $47.4$ & $69.1$ & $\mathbf{64.8}$ & $32.8$ & $53.8$ & $\underline{37.0}$ & $\underline{88.1}$ & $58.2$ & $56.4$ \\
SWA & $743$M & $\mathbf{49.6}$ & $68.7$ & $63.5$ & $31.6$ & $\underline{55.1}$ & $34.4$ & $85.2$ & $\underline{61.8}$ & $56.2$ \\
\textbf{FS-B} & $743$M & $\mathbf{49.6}$ & $\underline{70.0}$ & $63.8$ & $31.7$ & $\mathbf{56.0}$ & $33.4$ & $85.9$ & $59.1$ & $56.2$ \\
NSA & $1128$M & $47.4$ & $68.1$ & $63.8$ & $31.7$ & $54.3$ & $32.4$ & $\mathbf{88.8}$ & $61.4$ & $56.0$ \\
\textbf{FS-L} & $743$M & $48.1$ & $69.0$ & $62.6$ & $31.8$ & $53.6$ & $33.6$ & $85.4$ & $57.5$ & $55.2$ \\
NAtS-L & $793$M & $44.8$ & $68.2$ & $63.0$ & $29.0$ & $52.8$ & $\mathbf{37.4}$ & $85.3$ & $59.2$ & $55.0$ \\
Raven & $726$M & $45.8$ & $68.7$ & $63.4$ & $31.7$ & $54.0$ & $35.8$ & $84.4$ & $53.2$ & $54.6$ \\
\bottomrule
\end{tabular}
\par
\end{table}

\subsection{Empirical scaling}
\label{sec:efficiency}
To verify linear scaling empirically, we increase context from $4$k to $256$k
on one A100-40GB. FS-L's attention buffers remain constant at $766$ MiB for
the $340$M model and $1149$ MiB for the $800$M model. CUDA-graph decoding
runs at about $89$ and $75$ tokens/s, respectively, including retention
updates. Prefill throughput stays near $2.1$--$2.2$K tokens/s, so processing
time grows approximately linearly with input length.

These timings use bf16 parameters; Appendix~\ref{app:bounded-cache} compares
precision and execution paths.

The masked implementation prefills at $34.6$--$44.2$K tokens/s but keeps a growing KV cache.
Training at $340$M/$4$k is also slower than full attention: $98.4$K versus
$303$K tokens/s across eight GPUs; faster kernels for pruning, gathering, and
sparse readout are future work (Appendix~\ref{app:efficiency}).
Reducing $P$ to $16$ cuts $340$M attention storage to $320$ MiB with similar
decode speed, but accuracy varies non-monotonically with $P$
(Appendix~\ref{app:compression-rate}).

\section{Length extrapolation and learned retention}
\label{sec:analysis}
We analyze length generalization through an information-bottleneck view:
FS routes through cell summaries, and FS-L also restricts memory before
queries arrive (Appendix~\ref{app:analysis}).

\paragraph{Position dependence.}
Position handling affects how attention transfers to longer
sequences~\citep{press2022alibi,kazemnejad2023nope,chi2022dissecting}.
FS uses RoPE only within trained local offsets and NoPE for long-range token
attention, so longer contexts introduce no untrained rotary offsets. The erf
prior and local RoPE depend only on relative offsets, so shifting tokens, cells
and query blocks together preserves the operator at fixed content vectors
(Appendix~\ref{app:position-gate}). Added tokens can still introduce competing cells.

\paragraph{The information bottleneck in FS.}
The router identifies useful cells from their mean keys and maximum key norms.
Token attention then reads the selected cells' original keys and values.
We hypothesize that routing through these summaries encourages recognizable keypoints.

Each head reads at most $Ks$ token values, regardless of prefix length.
RMSNorm rescales the readout but cannot remove interference from irrelevant
values; sparse selection controls which values can contribute.
On a shared $256$k prompt, both variants at both scales select the needle
cell over $166{,}000$ tokens away (Appendix~\ref{app:generalization}).

\paragraph{The additional bottleneck in FS-L.}
FS-L discards cells before future queries arrive. This can tolerate growing
background when useful information remains concentrated in the retained
state, but not dense, independently queryable facts (Section~\ref{sec:fsb}).

CPT adapts the model to this restriction. At $340$M/$4$k, it recovers
S-NIAH-1 from $62$ under immediate pruning to $76$--$90$ across runs ($90$ for the
evaluated FS-L); unpruned continuations
retain $100$ (Appendix~\ref{app:sparse-fail}). FS-L retains $14.3\%$ of eligible
cells at $340$M/$16$k; Appendix~\ref{app:routing-probe} separates retention,
selection, and answering.

\paragraph{Budgets and ablations.}
$P$ sets the high-norm cells retained per head; $K$ sets the cells read per
query and head. Larger $K$ cannot recover a pruned cell; larger $P$ expands
availability but need not improve answers. We test adapted budgets and
retrieval among several independently queryable facts
(Appendix~\ref{app:compression-rate}).

The local NoPE control changes both rotation and local bias. It preserves
S-NIAH-1 extrapolation but reduces downstream accuracy from $51.9$ to $38.3$.
Removing the long-range norm-weighted erf term retains strong S-NIAH-1 recall, worsens
S-NIAH-2, and improves S-NIAH-3 (Appendix~\ref{app:ablations}). Since
$g_h\ge\erf(2)\simeq0.9953$ for eligible cells, the term acts mainly as a norm
bonus, so we do not credit extrapolation to its distance dependence
(Appendix~\ref{app:position-gate}).

\section{Related Work}
\label{sec:related}

\paragraph{Selective recurrent memory.}
Mamba, Mamba-2, and GDN compress history in recurrent
states~\citep{gu2023mamba,dao2024mamba2,yang2024gatedeltanet};
GSA and Raven use softmax over memory slots~\citep{zhang2024gsa,afzal2026raven}.
FS preserves token keys and values; summaries locate evidence.
Mixture-of-Experts and Mixture-of-Memories route to parameter experts or
recurrent states~\citep{shazeer2017moe,du2025mom}; FS routes to token cells.

\paragraph{Native block-sparse attention.}
MoBA~\citep{lu2025moba} also routes by mean keys and reads selected tokens,
always including the causal current block. NSA~\citep{yuan2025nsa} learns
compressed keys and values, derives selection scores from compressed attention,
and includes compressed values in its readout. FS adds norm priorities and
local-only RoPE to mean-key routing and selected-token readout. Training scores
all eligible cells but reads only selected values. FS-L prunes before query
matching; FS-B searches key hierarchies over the archive. Both reuse learned
token keys across memory and search budgets without a learned chunk compressor.

\paragraph{Sparse token retrieval.}
Landmark Attention uses learned landmarks~\citep{mohtashami2023landmark};
InfLLM uses representative tokens~\citep{xiao2024infllm}.
DRT~\citep{hu2025gca} and RAMba~\citep{hu2025hsa} learn chunk retrieval and
extrapolate beyond training length; routing scores weight their chunk outputs.
FS hard-selects by fixed mean/norm summaries, then applies one softmax over
selected tokens. Keys learn through token readout and local attention, without
gradients through selection indices (Appendix~\ref{app:retrieval-comparison}).

\paragraph{KV-cache selection and pruning.}
QUEST selects pages using key bounds~\citep{tang2024quest}.
H$_2$O and SnapKV prune using observed attention~\citep{zhang2023h2o,li2024snapkv};
\citet{devoto2024norm} retain low-norm keys. FS-L retains high-norm cells and
adapts representations to that fixed rule through CPT.
Retention needs neither a future query nor accumulated attention scores.
NAtS-L combines retained-token attention with GDN~\citep{deng2026natsl}; its
softmax history can grow. StreamingLLM keeps sinks and recent
tokens~\citep{xiao2024streamingllm}; FS-L can retain cells from anywhere in the prefix.

\paragraph{Sparsity and length generalization.}
ASEntmax studies sparse weights and extrapolation~\citep{vasylenko2026sparse};
FS uses softmax over selected tokens. Position interpolation and YaRN rescale
RoPE~\citep{chen2023pi,peng2024yarn}. RoPE/NoPE
hybrids~\citep{yang2025ropenope,puvvada2025swan} combine local RoPE and global NoPE
layers; Kimi Linear and Kimi K3 combine recurrence with global NoPE and train on
long contexts~\citep{kimiteam2025kda,kimiteam2026k3}. FS couples local RoPE with
sparse NoPE retrieval within each head, training at short context throughout.
Its long-range branch never encounters unseen rotary offsets.

\section{Limitations}
\label{sec:limitations}
Experiments stop at the $800$M scale. FS-L's fixed retention budget can discard
evidence before a query arrives. Its current CPT recovers S-NIAH-1 lost to
immediate pruning, but the resulting $800$M model has a lower RULER mean than
base FS at every evaluated length (Table~\ref{tab:matched-cpt}). Finding a CPT
objective that improves the suite mean within FS-L's memory budget remains open.
Retrieval-weighted CPT, as used for FS-B, motivates this direction
(Appendix~\ref{app:fsb-cpt}).
FS-B keeps growing memory and improves the RULER mean, but trails Raven-full
on $800$M multi-query and multi-value retrieval (Table~\ref{tab:ruler-full}).
Learning its search policy, rather than using a fixed inference algorithm,
is another direction for future work.

\section{Conclusion}
\label{sec:conclusion}
FreeSolo's central thesis is that memory summaries need only locate evidence;
the evidence can remain in original token values. Mean summaries of learned
token keys support hierarchical search, maximum token-key norms support pruning,
and local-only RoPE keeps distant retrieval free of unseen rotary offsets.
FS-L and FS-B show how this representation supports bounded memory or a growing
archive after short CPT, with zero-shot retrieval at $16$k and single-needle
recall at $256$k from $4$k training. This separates learning useful memory keys
from choosing the capacity, search effort, and read bandwidth of the memory.

\referencespagebreak
\bibliography{references}

\clearpage
\appendix

\section{Hyperparameters and experimental settings}
\label{app:hyperparams}

\paragraph{Architecture.}
FreeSolo uses LLaMA-style decoders with RMSNorm, SwiGLU, and a $32$k LLaMA
tokenizer. Both scales have $24$ layers and $16$ heads. The $341$M model has
width $1024$ and head width $64$; the $743$M model has width $1536$ and head
width $96$. Attention uses $w=512$, $B=128$, $s=32$, $K=4$, $\sigma_h=4$,
and $\lambda_h=1$. FS-L retains $P=64$ norm-ranked cells and $W=8$ recent
cells.

\paragraph{Pretraining.}
We train at $4096$ tokens in bf16 with AdamW, cosine learning-rate decay,
weight decay $0.1$, gradient clipping at $1.0$, and $524{,}288$ tokens per step.
The $341$M FS run and FS-L from-scratch control use peak learning rate
$3\times10^{-4}$, floor $3\times10^{-5}$, and $1000$ warmup steps.
The $743$M FS run uses $4\times10^{-4}$, $4\times10^{-5}$, and $2000$ warmup
steps. The base FS checkpoints are at $28{,}000$ steps for $341$M
and $60{,}000$ steps for $743$M, corresponding to $14{,}680{,}064{,}000$ and
$31{,}457{,}280{,}000$ training tokens. We refer to these scales by the
rounded $15$B/$32$B budgets. The smaller run makes about $2.2$ passes over
a $6.7$B-token FineWeb-Edu subset; the larger run uses less than one pass
over a $39$B-token corpus.

\paragraph{Continued pretraining.}
FS-L starts from FS with a fresh AdamW optimizer and pruning enabled.
The learning rate warms up to $4\times10^{-4}$ and decays toward
$4\times10^{-5}$. At $341$M, the schedule warms up for $150$ steps with a
$1500$-step cosine horizon, and we use the step-$1400$ checkpoint, selected on
S-NIAH-1 ($100$ examples); the $1500$-step continuations in
Appendix~\ref{app:sparse-fail} show the run-to-run range. At $743$M,
it warms up for $300$ steps and we use the final step-$3000$ checkpoint. These
correspond to $734{,}003{,}200$ and $1{,}572{,}864{,}000$ additional tokens. Both use the same $4096$-token
context, tokenizer, and corpus as FS; restarting the data order revisits
pretraining tokens.

\paragraph{RULER evaluation.}
\label{app:retrieval-setup}
We use NVIDIA RULER at commit \texttt{c3f5e3b4}, the LLaMA tokenizer,
\texttt{base} prompt template, seed $42$, and $100$ examples per task and length.
Generation caps are $128$ tokens for NIAH, $30$ for variable tracking, $120$
for common-word extraction, $50$ for frequent-word extraction, and $32$ for QA.
Decoding is greedy with EOS stopping. Non-QA tasks use \texttt{string\_match\_all}
and QA uses \texttt{string\_match\_part}, on a $0$--$100$ scale.
FS-L runs its bounded cache ($P=64$, $W=8$) with CUDA-graph decoding, FP32
parameters and BF16 autocast on A100 GPUs with $40$\,GB memory.
MoBA and NSA recompute the full forward pass for each generated token.
Lengths $2$k through $64$k denote $2048$ through $65{,}536$ tokens; table labels
$131$k and $256$k denote $131{,}072$ and $262{,}144$ tokens.

\paragraph{Downstream evaluation.}
The eight zero-shot tasks are HellaSwag, PIQA, ARC-easy, ARC-challenge,
WinoGrande, OpenBookQA, SciQ, and BoolQ, evaluated with the LM Evaluation
Harness.

\paragraph{Baselines.}
At $340$M, NAtS-L uses $21$ layers and $64$-token routing blocks.
Table~\ref{tab:baseline-budgets} lists the evaluated pretraining checkpoint of
every model; all runs use $4096$-token sequences. Raven is standalone at both scales,
with $21$ layers, four heads, $256$ memory slots, top-$32$ writes, and
\texttt{gate\_logit\_normalizer}${=}8$; its $726$M model has width $1536$.
Raven-full is the variant with full NoPE attention (Section~\ref{sec:setup}).
TF++ uses full RoPE attention, SWA a $512$-token RoPE window, and InfLLM
the TF++ checkpoint. NSA uses the paper's configuration (compression blocks of
$32$ tokens with stride $16$, $16$ selected $64$-token blocks, a $512$-token
window, and four key/value heads). Its larger parameter counts come from the
key and value token-compression MLPs, which hold $208$M parameters at $340$M
and $467$M at $800$M; NSA is $172$M and $385$M larger than the matching MoBA
or FreeSolo backbone.
For $800$M GDN, RULER uses \texttt{fla} $0.5.2$ and \texttt{transformers} $5.14.1$;
downstream tasks use versions $0.3.2$ and $4.56.1$, respectively, with
LM Evaluation Harness $0.4.13$.

\begin{table}[ht]
\centering
\caption{\textbf{Evaluated pretraining checkpoints.} Steps and training tokens
at $524{,}288$ tokens per step. MoBA is trained only at the smaller scale.
FS-L and FS-B continue from the FS checkpoints: FS-L adds $0.73$B ($340$M)
and $1.57$B ($800$M) CPT tokens; FS-B uses the FS-L weights at $340$M and
$0.42$B retrieval-weighted CPT tokens at $800$M (Appendix~\ref{app:fsb-cpt}).}
\label{tab:baseline-budgets}
\small\setlength{\tabcolsep}{6pt}
\begin{tabular}{@{}lrrrr@{}}
\toprule
& \multicolumn{2}{c}{$340$M} & \multicolumn{2}{c}{$800$M} \\
\cmidrule(lr){2-3}\cmidrule(l){4-5}
Model & Steps & Tokens & Steps & Tokens \\
\midrule
FS & $28{,}000$ & $14.68$B & $60{,}000$ & $31.46$B \\
TF++ & $28{,}610$ & $15.00$B & $60{,}000$ & $31.46$B \\
SWA & $28{,}000$ & $14.68$B & $60{,}000$ & $31.46$B \\
MoBA & $28{,}000$ & $14.68$B & -- & -- \\
NSA & $28{,}000$ & $14.68$B & $57{,}500$ & $30.15$B \\
GDN & $28{,}610$ & $15.00$B & $61{,}035$ & $32.00$B \\
NAtS-L & $28{,}610$ & $15.00$B & $61{,}035$ & $32.00$B \\
Raven & $28{,}610$ & $15.00$B & $61{,}035$ & $32.00$B \\
Raven-full & $28{,}610$ & $15.00$B & $61{,}035$ & $32.00$B \\
\bottomrule
\end{tabular}
\end{table}

\FloatBarrier

\section{Full RULER task breakdown}
\label{app:ruler-full}
Table~\ref{tab:ruler-full} reports individual task scores.
FS and FS-L improve long-context retrieval, while multi-key retrieval, variable
tracking, and word aggregation remain difficult. Raven retains strong S-NIAH-1
recall at $800$M, while FS-L performs better on S-NIAH-2/3 and the suite mean.
FS-B has the highest super-linear suite means from $4$k on at both scales
(Table~\ref{tab:ruler}) but remains below Raven-full on $800$M MQ- and MV-NIAH.

MK, MQ, and MV denote multi-key, multi-query, and multi-value NIAH; VT denotes
variable tracking; CWE and FWE denote common- and frequent-word extraction.
QA-2 is unavailable at $2$k/$4$k, so the mean covers $12$ tasks there and $13$
at $8$k/$16$k; ties share a mark.
NAtS-L's linear-time grouping assumes a bounded retained-token count
(Section~\ref{sec:setup}).

\begin{table}[ht]
\caption{\textbf{RULER results by task} ($0$--$100$) at $340$M and $800$M.
Bold and underline mark the best and second-best scores within each group (super-linear or linear cost) and scale.
Dashes denote unavailable results.}
\label{tab:ruler-full}
\begin{center}
\footnotesize\setlength{\tabcolsep}{2.5pt}
\resizebox{\textwidth}{!}{\begin{tabular}{@{}l cccc cccc cccc|cccc cccc cccc@{}}
\toprule
 & \multicolumn{12}{c|}{Super-linear} & \multicolumn{12}{c}{Linear} \\
\cmidrule(lr){2-13}\cmidrule(lr){14-25}
 & \multicolumn{4}{c}{Raven-full} & \multicolumn{4}{c}{\textbf{FS}} & \multicolumn{4}{c|}{\textbf{FS-B}} & \multicolumn{4}{c}{\textbf{FS-L}} & \multicolumn{4}{c}{NAtS-L} & \multicolumn{4}{c}{Raven} \\
\cmidrule(lr){2-5}\cmidrule(lr){6-9}\cmidrule(lr){10-13}\cmidrule(lr){14-17}\cmidrule(lr){18-21}\cmidrule(lr){22-25}
Task & $2$k & $4$k & $8$k & $16$k & $2$k & $4$k & $8$k & $16$k & $2$k & $4$k & $8$k & $16$k & $2$k & $4$k & $8$k & $16$k & $2$k & $4$k & $8$k & $16$k & $2$k & $4$k & $8$k & $16$k \\
\midrule
\multicolumn{25}{@{}l}{\emph{$\approx$340M scale}} \\
\midrule
S-NIAH-1 & $\mathbf{100}$ & $\mathbf{99}$ & $83$ & $39$ & $\underline{99}$ & $\mathbf{99}$ & $\mathbf{100}$ & $\mathbf{100}$ & $\mathbf{100}$ & $\mathbf{99}$ & $\underline{99}$ & $\underline{99}$ & $\underline{99}$ & $88$ & $\underline{90}$ & $\mathbf{87}$ & $\mathbf{100}$ & $\underline{94}$ & $37$ & $19$ & $95$ & $\mathbf{96}$ & $\mathbf{93}$ & $\underline{81}$ \\
S-NIAH-2 & $\mathbf{100}$ & $\underline{89}$ & $\underline{41}$ & $7$ & $\mathbf{100}$ & $77$ & $\underline{41}$ & $\underline{17}$ & $\mathbf{100}$ & $\mathbf{99}$ & $\mathbf{89}$ & $\mathbf{60}$ & $\mathbf{99}$ & $\mathbf{70}$ & $\underline{34}$ & $\mathbf{20}$ & $89$ & $\underline{22}$ & $20$ & $\underline{3}$ & $\underline{92}$ & $\mathbf{70}$ & $\mathbf{36}$ & $\underline{3}$ \\
S-NIAH-3 & $38$ & $1$ & $0$ & $\underline{0}$ & $\underline{58}$ & $\underline{14}$ & $\underline{7}$ & $\underline{0}$ & $\mathbf{91}$ & $\mathbf{78}$ & $\mathbf{64}$ & $\mathbf{27}$ & $\mathbf{70}$ & $\mathbf{17}$ & $\mathbf{9}$ & $\mathbf{1}$ & $\underline{36}$ & $\underline{6}$ & $\underline{1}$ & $\underline{0}$ & $3$ & $0$ & $0$ & $\underline{0}$ \\
MK-NIAH-1 & $\underline{25}$ & $\underline{22}$ & $\underline{18}$ & $\underline{18}$ & $\mathbf{27}$ & $20$ & $15$ & $5$ & $\mathbf{27}$ & $\mathbf{25}$ & $\mathbf{19}$ & $\mathbf{24}$ & $\mathbf{31}$ & $\mathbf{24}$ & $\mathbf{21}$ & $\mathbf{10}$ & $\underline{17}$ & $\underline{18}$ & $\underline{10}$ & $\underline{4}$ & $11$ & $11$ & $\underline{10}$ & $3$ \\
MK-NIAH-2 & $0$ & $\mathbf{1}$ & $\mathbf{0}$ & $\underline{0}$ & $\underline{1}$ & $\underline{0}$ & $\mathbf{0}$ & $\underline{0}$ & $\mathbf{2}$ & $\underline{0}$ & $\mathbf{0}$ & $\mathbf{1}$ & $\mathbf{0}$ & $\mathbf{1}$ & $\mathbf{0}$ & $\mathbf{0}$ & $\mathbf{0}$ & $\underline{0}$ & $\mathbf{0}$ & $\mathbf{0}$ & $\mathbf{0}$ & $\underline{0}$ & $\mathbf{0}$ & $\mathbf{0}$ \\
MK-NIAH-3 & $0$ & $\mathbf{0}$ & $\mathbf{0}$ & $\underline{0}$ & $\underline{1}$ & $\mathbf{0}$ & $\mathbf{0}$ & $\mathbf{1}$ & $\mathbf{2}$ & $\mathbf{0}$ & $\mathbf{0}$ & $\underline{0}$ & $\mathbf{0}$ & $\mathbf{0}$ & $\mathbf{0}$ & $\mathbf{0}$ & $\mathbf{0}$ & $\mathbf{0}$ & $\mathbf{0}$ & $\mathbf{0}$ & $\mathbf{0}$ & $\mathbf{0}$ & $\mathbf{0}$ & $\mathbf{0}$ \\
MQ-NIAH & $\underline{20.8}$ & $13.3$ & $6.8$ & $\underline{5.0}$ & $20.2$ & $\underline{20.8}$ & $\underline{12.8}$ & $2.8$ & $\mathbf{23.8}$ & $\mathbf{22.5}$ & $\mathbf{18.5}$ & $\mathbf{12.5}$ & $\mathbf{21.8}$ & $\mathbf{21.5}$ & $\underline{10.5}$ & $\mathbf{3.8}$ & $\underline{17.5}$ & $15.2$ & $4.8$ & $2.8$ & $15.0$ & $\underline{16.5}$ & $\mathbf{13.2}$ & $\underline{3.0}$ \\
MV-NIAH & $19.3$ & $17.8$ & $6.8$ & $6.5$ & $\underline{23.2}$ & $\underline{18.5}$ & $\underline{15.5}$ & $\underline{9.8}$ & $\mathbf{24.5}$ & $\mathbf{22.0}$ & $\mathbf{23.5}$ & $\mathbf{21.0}$ & $\mathbf{24.3}$ & $\mathbf{19.8}$ & $\mathbf{16.3}$ & $\mathbf{7.3}$ & $\underline{18.8}$ & $8.8$ & $4.5$ & $\underline{7.0}$ & $11.2$ & $\underline{12.0}$ & $\underline{12.0}$ & $4.5$ \\
VT & $0.8$ & $\mathbf{1.4}$ & $0.4$ & $0.2$ & $\underline{2.2}$ & $\underline{1.2}$ & $\underline{0.6}$ & $\mathbf{1.6}$ & $\mathbf{3.2}$ & $\underline{1.2}$ & $\mathbf{1.0}$ & $\underline{1.0}$ & $\underline{2.8}$ & $0.0$ & $\underline{0.0}$ & $\underline{0.0}$ & $\mathbf{4.8}$ & $\underline{0.6}$ & $\underline{0.0}$ & $\underline{0.0}$ & $2.6$ & $\mathbf{2.0}$ & $\mathbf{1.4}$ & $\mathbf{1.0}$ \\
CWE & $0.2$ & $0.6$ & $0.0$ & $0.1$ & $\underline{4.1}$ & $\underline{9.2}$ & $\underline{5.2}$ & $\underline{0.8}$ & $\mathbf{5.2}$ & $\mathbf{10.3}$ & $\mathbf{6.5}$ & $\mathbf{1.8}$ & $\mathbf{4.9}$ & $\mathbf{9.1}$ & $\mathbf{4.0}$ & $\mathbf{0.4}$ & $\underline{2.6}$ & $\underline{5.0}$ & $\underline{0.2}$ & $\underline{0.3}$ & $1.0$ & $0.2$ & $\underline{0.2}$ & $0.1$ \\
FWE & $0.0$ & $0.0$ & $0.0$ & $0.0$ & $\underline{8.7}$ & $\underline{3.3}$ & $\underline{5.0}$ & $\underline{3.3}$ & $\mathbf{9.0}$ & $\mathbf{6.0}$ & $\mathbf{11.7}$ & $\mathbf{5.0}$ & $\mathbf{9.7}$ & $\underline{1.0}$ & $\mathbf{2.3}$ & $\mathbf{2.7}$ & $\underline{3.0}$ & $\mathbf{1.7}$ & $\underline{0.3}$ & $\underline{2.3}$ & $0.0$ & $0.0$ & $0.0$ & $0.0$ \\
QA-1 & $16$ & $\underline{13}$ & $\mathbf{9}$ & $\mathbf{7}$ & $\mathbf{25}$ & $\mathbf{15}$ & $\underline{5}$ & $\mathbf{7}$ & $\underline{23}$ & $10$ & $4$ & $\underline{4}$ & $\mathbf{26}$ & $\underline{8}$ & $\underline{4}$ & $3$ & $\underline{19}$ & $\mathbf{12}$ & $\mathbf{9}$ & $\underline{4}$ & $15$ & $6$ & $\underline{4}$ & $\mathbf{6}$ \\
QA-2 & -- & -- & $\mathbf{10}$ & $\mathbf{10}$ & -- & -- & $8$ & $5$ & -- & -- & $\underline{9}$ & $\underline{6}$ & -- & -- & $\underline{10}$ & $4$ & -- & -- & $\underline{10}$ & $\mathbf{10}$ & -- & -- & $\mathbf{12}$ & $\underline{9}$ \\
\midrule
Available-task mean & $26.7$ & $21.5$ & $13.5$ & $7.1$ & $\underline{30.8}$ & $\underline{23.2}$ & $\underline{16.5}$ & $\underline{11.8}$ & $\mathbf{34.2}$ & $\mathbf{31.1}$ & $\mathbf{26.6}$ & $\mathbf{20.2}$ & $\mathbf{32.4}$ & $\mathbf{21.6}$ & $\mathbf{15.5}$ & $\mathbf{10.7}$ & $\underline{25.6}$ & $15.3$ & $7.4$ & $4.0$ & $20.5$ & $\underline{17.8}$ & $\underline{14.0}$ & $\underline{8.5}$ \\
\midrule
\multicolumn{25}{@{}l}{\emph{$\approx$800M scale}} \\
\midrule
S-NIAH-1 & $\mathbf{100}$ & $\mathbf{100}$ & $\mathbf{100}$ & $\mathbf{100}$ & $\underline{97}$ & $91$ & $93$ & $\underline{95}$ & $\underline{97}$ & $\underline{98}$ & $\underline{97}$ & $\mathbf{100}$ & $\mathbf{100}$ & $\underline{92}$ & $\underline{84}$ & $\underline{84}$ & $\underline{86}$ & $72$ & $59$ & $21$ & $\mathbf{100}$ & $\mathbf{100}$ & $\mathbf{100}$ & $\mathbf{100}$ \\
S-NIAH-2 & $\mathbf{100}$ & $88$ & $81$ & $69$ & $\mathbf{100}$ & $\underline{99}$ & $\underline{92}$ & $\underline{72}$ & $\mathbf{100}$ & $\mathbf{100}$ & $\mathbf{95}$ & $\mathbf{88}$ & $\underline{99}$ & $\mathbf{99}$ & $\mathbf{89}$ & $\mathbf{81}$ & $\mathbf{100}$ & $76$ & $28$ & $3$ & $\underline{99}$ & $\underline{97}$ & $\underline{81}$ & $\underline{26}$ \\
S-NIAH-3 & $\underline{91}$ & $58$ & $21$ & $4$ & $85$ & $\underline{71}$ & $\underline{51}$ & $\underline{34}$ & $\mathbf{96}$ & $\mathbf{93}$ & $\mathbf{87}$ & $\mathbf{84}$ & $\mathbf{90}$ & $\mathbf{71}$ & $\mathbf{55}$ & $\mathbf{40}$ & $\underline{62}$ & $\underline{24}$ & $\underline{6}$ & $\underline{0}$ & $32$ & $7$ & $1$ & $\underline{0}$ \\
MK-NIAH-1 & $37$ & $\mathbf{36}$ & $\underline{29}$ & $\mathbf{32}$ & $\mathbf{53}$ & $32$ & $\mathbf{30}$ & $28$ & $\underline{50}$ & $\underline{34}$ & $28$ & $\underline{31}$ & $\mathbf{43}$ & $\mathbf{32}$ & $\mathbf{27}$ & $\mathbf{23}$ & $\underline{31}$ & $\underline{20}$ & $7$ & $\underline{3}$ & $30$ & $19$ & $\underline{14}$ & $\underline{3}$ \\
MK-NIAH-2 & $3$ & $\underline{1}$ & $\underline{0}$ & $0$ & $\underline{7}$ & $0$ & $\mathbf{1}$ & $\underline{1}$ & $\mathbf{9}$ & $\mathbf{4}$ & $\mathbf{1}$ & $\mathbf{2}$ & $\mathbf{8}$ & $\mathbf{3}$ & $\mathbf{1}$ & $\mathbf{1}$ & $0$ & $\underline{0}$ & $\underline{0}$ & $\underline{0}$ & $\underline{3}$ & $\underline{0}$ & $\underline{0}$ & $\underline{0}$ \\
MK-NIAH-3 & $\mathbf{3}$ & $\mathbf{2}$ & $\mathbf{0}$ & $\mathbf{0}$ & $0$ & $0$ & $\mathbf{0}$ & $\mathbf{0}$ & $\underline{2}$ & $\underline{1}$ & $\mathbf{0}$ & $\mathbf{0}$ & $\mathbf{2}$ & $\mathbf{0}$ & $\mathbf{1}$ & $\mathbf{0}$ & $\underline{0}$ & $\mathbf{0}$ & $\underline{0}$ & $\mathbf{0}$ & $\underline{0}$ & $\mathbf{0}$ & $\underline{0}$ & $\mathbf{0}$ \\
MQ-NIAH & $\mathbf{69.5}$ & $\mathbf{46.5}$ & $\mathbf{44.0}$ & $\mathbf{38.3}$ & $34.5$ & $25.8$ & $22.8$ & $23.0$ & $\underline{45.8}$ & $\underline{28.3}$ & $\underline{28.8}$ & $\underline{27.3}$ & $\underline{26.2}$ & $\mathbf{22.5}$ & $\mathbf{16.0}$ & $\mathbf{15.5}$ & $16.5$ & $11.5$ & $1.5$ & $0.3$ & $\mathbf{27.8}$ & $\underline{19.8}$ & $\underline{11.0}$ & $\underline{7.2}$ \\
MV-NIAH & $\mathbf{56.3}$ & $\mathbf{41.3}$ & $\mathbf{39.0}$ & $\mathbf{39.0}$ & $28.0$ & $26.3$ & $22.8$ & $18.8$ & $\underline{38.3}$ & $\underline{30.3}$ & $\underline{29.8}$ & $\underline{26.3}$ & $\mathbf{26.2}$ & $\mathbf{20.8}$ & $\mathbf{21.0}$ & $\mathbf{17.0}$ & $13.5$ & $10.8$ & $3.0$ & $0.5$ & $\underline{25.8}$ & $\underline{17.0}$ & $\underline{13.5}$ & $\underline{9.0}$ \\
VT & $0.6$ & $\underline{0.2}$ & $\underline{0.6}$ & $\mathbf{0.6}$ & $\underline{1.6}$ & $\underline{0.2}$ & $0.4$ & $\underline{0.0}$ & $\mathbf{3.8}$ & $\mathbf{1.4}$ & $\mathbf{1.0}$ & $\mathbf{0.6}$ & $\mathbf{3.8}$ & $\mathbf{0.8}$ & $\underline{0.4}$ & $\underline{0.6}$ & $0.0$ & $\underline{0.2}$ & $0.0$ & $0.0$ & $\underline{1.6}$ & $\mathbf{0.8}$ & $\mathbf{0.6}$ & $\mathbf{1.0}$ \\
CWE & $3.3$ & $9.3$ & $5.4$ & $1.2$ & $\mathbf{8.4}$ & $\mathbf{26.7}$ & $\mathbf{13.5}$ & $\underline{3.3}$ & $\underline{7.8}$ & $\underline{26.4}$ & $\underline{8.9}$ & $\mathbf{4.4}$ & $\mathbf{4.5}$ & $\mathbf{13.0}$ & $\mathbf{7.1}$ & $\mathbf{1.2}$ & $0.4$ & $0.7$ & $0.0$ & $0.0$ & $\underline{1.9}$ & $\underline{1.3}$ & $\underline{0.1}$ & $\underline{0.1}$ \\
FWE & $1.7$ & $1.7$ & $0.0$ & $0.0$ & $\underline{16.7}$ & $\underline{6.0}$ & $\underline{5.0}$ & $\underline{5.7}$ & $\mathbf{17.7}$ & $\mathbf{8.7}$ & $\mathbf{7.0}$ & $\mathbf{8.0}$ & $0.3$ & $0.0$ & $\underline{1.7}$ & $\underline{0.0}$ & $\underline{1.7}$ & $\underline{3.3}$ & $\mathbf{4.7}$ & $\underline{0.0}$ & $\mathbf{18.0}$ & $\mathbf{11.0}$ & $1.3$ & $\mathbf{3.7}$ \\
QA-1 & $32$ & $20$ & $10$ & $\mathbf{12}$ & $\mathbf{42}$ & $\mathbf{30}$ & $\underline{12}$ & $9$ & $\underline{38}$ & $\underline{27}$ & $\mathbf{13}$ & $\underline{10}$ & $\mathbf{43}$ & $\mathbf{30}$ & $\mathbf{11}$ & $\mathbf{11}$ & $\underline{23}$ & $\underline{14}$ & $\mathbf{11}$ & $0$ & $22$ & $\underline{14}$ & $\underline{6}$ & $\underline{5}$ \\
QA-2 & -- & -- & $\underline{15}$ & $\underline{13}$ & -- & -- & $\underline{15}$ & $11$ & -- & -- & $\mathbf{17}$ & $\mathbf{14}$ & -- & -- & $\mathbf{19}$ & $\mathbf{16}$ & -- & -- & $\underline{12}$ & $9$ & -- & -- & $11$ & $\underline{11}$ \\
\midrule
Available-task mean & $\underline{41.4}$ & $33.7$ & $26.5$ & $\underline{23.8}$ & $39.4$ & $\underline{34.0}$ & $\underline{27.6}$ & $23.1$ & $\mathbf{42.1}$ & $\mathbf{37.7}$ & $\mathbf{31.8}$ & $\mathbf{30.4}$ & $\mathbf{37.2}$ & $\mathbf{32.0}$ & $\mathbf{25.6}$ & $\mathbf{22.3}$ & $27.8$ & $19.4$ & $10.2$ & $2.8$ & $\underline{30.1}$ & $\underline{23.9}$ & $\underline{18.4}$ & $\underline{12.8}$ \\
\bottomrule
\end{tabular}}
\end{center}
\end{table}

\FloatBarrier

\section{Efficiency benchmarks}
\label{app:efficiency}
The streaming implementation maintains fixed attention buffers and nearly
constant decode throughput from $4$k to $256$k. Absolute throughput still
lags optimized local and recurrent baselines. We measure training, prefill,
and decoding separately. These benchmarks use the
models' available optimized kernels; the results depend on batch size, software
stack, and execution mode as well as architecture.

\paragraph{Training.}
Table~\ref{tab:eff-train} reports aggregate forward-and-backward throughput on
eight A100-40GB GPUs, with $4096$-token sequences, bf16, and
\texttt{torch.compile}. All rows use batch $1$ per GPU. FS-B changes inference
only: it uses the FS-L weights at $340$M, and its CPT at $800$M trains with FS
selection, so it has no separate row. Raven and NAtS-L are
measured with the fla/flame (FSDP) harness (sharded parameters and optimizer
states), so their peak memory is not directly comparable and is omitted, and their
throughput carries some FSDP communication overhead. Bold and underline
mark the best and second-best results (higher throughput, lower memory).

GDN uses flash-linear-attention, NSA uses its
Triton kernel, MoBA uses SDPA, and FreeSolo uses fused indexed-cell attention.
FS, TF++, and SWA run on PyTorch $2.5$; the other models use PyTorch $2.9$.
Sparse long-range retrieval reduces scaling cost, but has overhead at the short
training context.

\begin{table}[ht]
\caption{\textbf{Training throughput and peak memory} at $340$M scale
on eight A100-40GB GPUs. Raven and NAtS-L use the fla/flame FSDP harness; see the text.}
\label{tab:eff-train}
\begin{center}
\small\setlength{\tabcolsep}{10pt}
\begin{tabular}{@{}lcc@{}}
\toprule
Model & Train tok/s & Peak mem (GiB) \\
\midrule
TF++ (full attention) & $\mathbf{303}$K & $\mathbf{15.0}$ \\
SWA (sliding-$512$) & $\underline{228}$K & $\underline{15.5}$ \\
GDN & $158$K & $21.6$ \\
MoBA & $140$K & $27.0$ \\
NSA & $28$K & $31.9$ \\
Raven & $140$K & -- \\
NAtS-L & $126$K & -- \\
\textbf{FS} & $99.3$K & $16.5$ \\
\textbf{FS-L} & $98.4$K & $16.6$ \\
\bottomrule
\end{tabular}
\end{center}
\end{table}

\paragraph{Inference.}
Table~\ref{tab:eff-infer} uses one A100-40GB, bf16, and batch $1$. Prefill is one
forward pass over the prompt; decode processes one new token using each model's
KV cache or recurrent state. All table entries use eager execution, without
\texttt{torch.compile} or CUDA-graph capture. TF++ uses FlashAttention-2~\citep{dao2023flash2}; FreeSolo
uses fused indexed-cell prefill and a fused sliding-window local branch.
The other baselines use their native attention or recurrent kernels, with an
incremental reference path for NSA decoding.

FS-L prefill stays nearly flat from $4$k to $131$k, where FS runs out of
memory, while several baselines decode faster. FS-B runs to $131$k within
$19$\,GiB, but the measured reference uses unfused forest construction,
traversal and reranking and repeatedly copies growing cache buffers
(Appendix~\ref{app:fsb-cost}). Its eager throughput ($1.1$--$1.5$K tokens/s prefill,
$5.3$--$6.1$ tokens/s decode) reflects this implementation rather than the
$O(N\log N)$ search; with CUDA-graph decode it reaches $19.1$--$19.8$ tokens/s. The benefit is sustained
long-context retrieval with bounded query scoring, rather than the highest
throughput at every length. Table~\ref{tab:eff-infer} uses an FS-L
implementation that stores all KV and rescans priorities
(Section~\ref{sec:complexity}); Table~\ref{tab:bounded-cache} measures the
streaming cache.

\begin{table}[ht]
\caption{\textbf{Prefill and eager decode throughput} at $340$M scale
on one A100-40GB (bf16, batch $1$). OOM denotes the $40$\,GB memory limit.}
\label{tab:eff-infer}
\begin{center}
\footnotesize\setlength{\tabcolsep}{4pt}
\begin{tabular}{@{}lcccccccc@{}}
\toprule
& \multicolumn{4}{c}{Prefill (Ktok/s)} & \multicolumn{4}{c}{Decode (tok/s)} \\
\cmidrule(lr){2-5}\cmidrule(lr){6-9}
Model & $4$k & $16$k & $64$k & $131$k & $4$k & $16$k & $64$k & $131$k \\
\midrule
TF++ (full attention) & $144.4$ & $123.8$ & $48.8$ & $26.2$ & $34.7$ & $34.7$ & $34.7$ & $20.5$ \\
SWA (sliding-$512$) & $143.5$ & $218.8$ & $248.6$ & $245.3$ & $33.1$ & $33.2$ & $33.2$ & $33.3$ \\
MoBA & $34.3$ & $48.6$ & $33.3$ & $20.2$ & $27.4$ & $26.9$ & $26.1$ & OOM \\
NSA & $87.0$ & $121.4$ & $89.8$ & $65.2$ & $15.8$ & $15.7$ & $15.8$ & $15.8$ \\
GDN & $65.7$ & $141.8$ & $146.8$ & $146.2$ & $25.5$ & $25.7$ & $25.6$ & $25.4$ \\
NAtS-L & $56.1$ & $132.6$ & $107.6$ & $83.7$ & $22.5$ & $22.7$ & $22.2$ & $22.9$ \\
Raven & $71.3$ & $137.7$ & $147.6$ & $148.9$ & $24.7$ & $24.8$ & $24.8$ & $24.8$ \\
FS & $30.9$ & $22.3$ & $10.9$ & OOM & $19.0$ & $18.7$ & $18.2$ & OOM \\
FS-L & $34.6$ & $43.6$ & $44.2$ & $39.6$ & $19.8$ & $19.9$ & $19.8$ & $19.8$ \\
FS-B & $1.5$ & $1.4$ & $1.2$ & $1.1$ & $6.1$ & $5.9$ & $5.6$ & $5.3$ \\
\bottomrule
\end{tabular}
\end{center}
\end{table}

\paragraph{Compilation and CUDA graphs.}
With \texttt{torch.compile}, FS-L prefill reaches $62.8$--$68.2$K tokens/s
from $4$k to $131$k, and CUDA-graph capture raises its decode to
$94.4$--$94.7$ tokens/s through $64$k; with the same capture, SWA, GDN, and
Raven decode at about $380$, $275$, and $262$ tokens/s.

\paragraph{Bounded cache.}
\label{app:bounded-cache}
The streaming implementation updates FS-L's top-$P$ pool exactly from newly
eligible cells. Let $\mathcal C$ be the eligible prefix cells and $\mathcal D$
those newly eligible at the next query-block boundary. Fixed priorities $a_c$
and a common deterministic tie rule give
\begin{equation}
  \operatorname{TopP}(\mathcal C\cup\mathcal D)
  =\operatorname{TopP}\!\left(\operatorname{TopP}(\mathcal C)\cup\mathcal D\right).
\end{equation}
A discarded cell already has at least $P$ cells ahead of it, so adding cells
cannot improve its rank. Once it also leaves the recent queue and local
window, its KV can be evicted.

The implementation keeps absolute cell indices for relative scores and
computes RoPE locally; position counters require $O(\log N)$ bits.
Prefill processes the transformer in chunks of
$2048$ tokens, with retention updates every $B=128$ tokens. Decode uses a
fixed local ring and compact cell buffers; pool updates occur outside the
CUDA graph and are included in the timing.

Table~\ref{tab:bounded-cache} uses both CPT checkpoints, bf16 parameters
and activations, and FP32 rotary frequencies on one A100-40GB. We report the
median of three runs after one warmup, with $257$ decode steps per run.
Attention-buffer bytes count distinct physical allocations across all heads
and layers, including retained, recent, local, pending, and graph input buffers.
The implementation keeps duplicate compact buffers, so this exceeds the size
of one packed retained set. Peak allocated GPU memory also includes parameters,
activations, and graph workspace; input/output token IDs are excluded from the
attention-buffer count.

\begin{table}[ht]
\caption{\textbf{Bounded FS-L inference.} Throughput uses CUDA-graph
decode; peak memory is the maximum across three trials.}
\label{tab:bounded-cache}
\begin{center}
\small\setlength{\tabcolsep}{6pt}
\begin{tabular}{@{}lrrrr@{}}
\toprule
Length & Attention buffers (MiB) & Peak (GiB) & Prefill (Ktok/s) & Decode (tok/s) \\
\midrule
\multicolumn{5}{@{}l}{\emph{$340$M scale}} \\
$4$k & $766.2$ & $1.71$ & $2.20$ & $89.3$ \\
$64$k & $766.2$ & $1.74$ & $2.21$ & $89.4$ \\
$256$k & $766.2$ & $1.77$ & $2.17$ & $88.3$ \\
\midrule
\multicolumn{5}{@{}l}{\emph{$800$M scale}} \\
$4$k & $1148.8$ & $2.96$ & $2.09$ & $75.2$ \\
$64$k & $1148.8$ & $2.98$ & $2.14$ & $75.1$ \\
$256$k & $1148.8$ & $3.02$ & $2.14$ & $74.1$ \\
\bottomrule
\end{tabular}
\end{center}
\end{table}

This establishes bounded attention storage with linear streaming work, at a
substantial prefill cost relative to Table~\ref{tab:eff-infer}. Both implementations
apply the same pruning rule, but precision and reduction order can change
generated answers. On $25$ prompts per S-NIAH task at $4$k and $16$k, bounded
execution at $340$M matches all $150$ correctness decisions of the masked
implementation (FP32 parameters, bf16 autocast). The bf16 CUDA-graph configuration used for
timing matches $60/60$ decisions at $340$M and $59/60$ at $800$M on ten prompts
per task and length, and $58/60$ for the adapted $P=16$ and $P=32$ checkpoints;
both scales retrieve all three tested $256$k S-NIAH-1 examples. FS-L results in
Tables~\ref{tab:ruler} and~\ref{tab:ruler-full} use the bounded cache on $100$
examples per cell, with FP32 parameters and bf16 autocast
(Appendix~\ref{app:retrieval-setup}). The FS-L continuation and adapted-budget
accuracies in Appendices~\ref{app:sparse-fail} and~\ref{app:learned-retention}
also use bounded execution. The routing probes and eligible-prompt answer
counts in Appendix~\ref{app:routing-probe} use the full-history masked
implementation, which applies the same pruning rule without evicting stored KV.

\clearpage

\section{Theory and ablation studies}
\label{app:analysis}
In this section, we present additional theoretical analysis, ablation studies,
and exploratory tests and visualizations (tagged \emph{Probe}).
The main findings are:

\paragraph{Ablations of short-range and long-range components (Appendix~\ref{app:ablations}).}
\begin{enumerate}
\item \emph{(Ablation)} Short-range RoPE is crucial for commonsense
accuracy; single-needle length extrapolation persists without it.
\item \emph{(Theory, Probe)} The long-range erf term acts mainly
as a key-norm bonus; its residual distance dependence rarely changes selection.
\end{enumerate}

\paragraph{Cell summaries and reading distance (Appendix~\ref{app:memory-access}).}
\begin{enumerate}
\item \emph{(Theory, Probe)} Cell summaries need only locate useful
evidence: the readout accesses the selected tokens' original values directly,
including needle reads over $166$k tokens away in the traced prompt.
\item \emph{(Theory)} Sparse readout limits interference from unrelated values;
readout normalization alone cannot undo their mixing.
\end{enumerate}

\paragraph{Norm pruning and retrieval coverage (Appendix~\ref{app:routing-probe}).}
\begin{enumerate}
\item \emph{(Probe)} Norm pruning substantially reduces
query-scoring work: FS-L retains about one-seventh of eligible cells at $16$k.
\item \emph{(Probe)} On harder retrieval tasks,
query-dependent selection remains a bottleneck even when the answer source
survives pruning.
\end{enumerate}

\paragraph{Continued pretraining (Appendix~\ref{app:sparse-fail}).}
\begin{enumerate}
\item \emph{(Ablation)} Short CPT helps FS-L adapt to the pruning
constraint, recovering much of the S-NIAH-1 accuracy lost when pruning is
applied without adaptation.
\item \emph{(Ablation)} Harder-task changes cannot be
attributed to pruning alone: they vary across seeds and also occur with
unpruned continuation.
\item \emph{(Ablation)} Retrieval-weighted CPT improves the RULER mean of
the full-prefix single-forest ablation (W32) at every evaluated length using only
short-context training.
\end{enumerate}

\paragraph{Retention budgets and multi-fact retrieval (Appendix~\ref{app:learned-retention}).}
\begin{enumerate}
\item \emph{(Theory)} Sustaining high accuracy when retrieving independently
queryable random facts requires storage to grow with their number.
\item \emph{(Ablation)} Adapted models do not improve
monotonically with retention capacity; a larger pool is not sufficient for
better retrieval.
\item \emph{(Probe)} Multi-fact failures extend beyond
pruning: unpruned and full-attention models also confuse values from the
same prefix.
\end{enumerate}

\subsection{Ablations of short-range and long-range components}
\label{app:ablations}\label{app:position-gate}
The ablations separate short-context quality from length extrapolation:
changing the local branch sharply affects downstream accuracy, while distant
retrieval persists. In long-range routing, the erf term acts mainly as a norm
bonus at $s=32$ and $\sigma_h=4$.

\paragraph{Position dependence.}
Local scores use $\mathbf q_i^\top \mathbf R(j-i)\mathbf k_j/\sqrt d$ with
$0\le i-j\le w$. Training at length $L>w$ covers every explicit rotary
offset used at evaluation. Long-range token scores are NoPE. The cell prior
$g_h(i-p_c)$ is unchanged by shifting query and cell coordinates together,
and approaches one with distance. At fixed content vectors and a consistent
tie rule, translating tokens, cell partitions, query blocks, and causal boundaries
together by a multiple of $B$ leaves the operator unchanged. These properties avoid new rotary offsets;
they do not make contextual representations invariant to added tokens.

\paragraph{Trained ablations.}
We train both controls with the $341$M FS recipe, $4$k context, and $15$B
training tokens. SR-NoPE (short range without RoPE) replaces local RoPE with
unrotated queries and keys and adds a $\log\erfc(\Delta/(2\sigma_h))$ bias with
$\Delta=i-j$, retaining the $512$-token window; it therefore changes both rotation
and local bias. LR-content-only keeps local RoPE and removes the entire
$\lambda_h a_c\erf((i-p_c)/(2\sigma_h))$ routing term, so long-range routing uses
the query--mean-key similarity alone.

\begin{table}[ht]
\caption{\textbf{RULER ablations at $340$M} ($0$--$100$).
Bold and underline mark the best and second-best scores, including ties.}
\label{tab:abl-ruler}
\begin{center}
\small\setlength{\tabcolsep}{3pt}
\begin{tabular}{@{}l ccccccc cccc cccc@{}}
\toprule
& \multicolumn{7}{c}{S-NIAH-1} & \multicolumn{4}{c}{S-NIAH-2} & \multicolumn{4}{c}{S-NIAH-3} \\
\cmidrule(lr){2-8}\cmidrule(lr){9-12}\cmidrule(lr){13-16}
Model & 2k & 4k & 8k & 16k & 32k & 64k & 131k & 2k & 4k & 8k & 16k & 2k & 4k & 8k & 16k \\
\midrule
\textbf{FS} & $\underline{99}$ & $\underline{99}$ & $\mathbf{100}$ & $\mathbf{100}$ & $\underline{98}$ & $\mathbf{100}$ & $\underline{98}$ & $\mathbf{100}$ & $\underline{77}$ & $\underline{41}$ & $\underline{17}$ & $\underline{58}$ & $14$ & $7$ & $\underline{0}$ \\
SR-NoPE & $\mathbf{100}$ & $\mathbf{100}$ & $\mathbf{100}$ & $\mathbf{100}$ & $\mathbf{100}$ & $\mathbf{100}$ & $\mathbf{100}$ & $\underline{98}$ & $\mathbf{90}$ & $\mathbf{71}$ & $\mathbf{29}$ & $47$ & $\underline{23}$ & $\mathbf{13}$ & $\underline{0}$ \\
LR-content-only & $95$ & $97$ & $\underline{96}$ & $\underline{97}$ & $95$ & $\underline{95}$ & $96$ & $80$ & $57$ & $38$ & $14$ & $\mathbf{71}$ & $\mathbf{25}$ & $\underline{11}$ & $\mathbf{1}$ \\
\bottomrule
\end{tabular}
\end{center}
\end{table}

\begin{table}[ht]
\caption{\textbf{Downstream accuracy of the ablations at $340$M} ($0$--$100$),
using the metrics in Table~\ref{tab:downstream-zs}.}
\label{tab:abl-cs}
\begin{center}
\small\setlength{\tabcolsep}{5pt}
\begin{tabular}{@{}lccccccccc@{}}
\toprule
Model & HellaS. & PIQA & ARC-e & ARC-c & WinoG. & OBQA & SciQ & BoolQ & Avg \\
\midrule
\textbf{FS} & $\underline{38.9}$ & $\mathbf{66.8}$ & $\mathbf{57.6}$ & $\underline{25.9}$ & $\underline{50.9}$ & $\mathbf{32.4}$ & $\underline{81.2}$ & $\mathbf{61.1}$ & $\underline{51.9}$ \\
SR-NoPE & $28.4$ & $55.7$ & $36.9$ & $23.7$ & $49.2$ & $25.6$ & $48.2$ & $38.7$ & $38.3$ \\
LR-content-only & $\mathbf{39.3}$ & $\underline{66.1}$ & $\underline{57.1}$ & $\mathbf{26.9}$ & $\mathbf{52.8}$ & $\underline{31.8}$ & $\mathbf{83.2}$ & $\underline{60.0}$ & $\mathbf{52.1}$ \\
\bottomrule
\end{tabular}
\end{center}
\end{table}

SR-NoPE preserves S-NIAH-1 extrapolation and improves S-NIAH-2, but lowers
average downstream accuracy from $51.9$ to $38.3$. The local FS configuration
therefore improves language-model quality without being required for single-needle
retrieval. LR-content-only retains $95$--$97$ on S-NIAH-1 and a downstream
average of $52.1$, while worsening S-NIAH-2 and improving S-NIAH-3. The norm
bonus is therefore not needed for single-needle extrapolation; it shifts
S-NIAH-2 and S-NIAH-3 in opposite directions.

\paragraph{A saturated norm prior.}
Every eligible cell satisfies $(c+1)s\le b_i\le i$, so its center
$p_c=(c+\tfrac12)s$ obeys $i-p_c\ge s/2$. Write the content term as
$\kappa_{ic}=\vb q_i^\top\bar{\vb k}_c/\sqrt d$. For $\lambda_h\ge0$, define
the constant-gate score $r^\infty_{ic}=\kappa_{ic}+\lambda_h a_c$. Then
\begin{equation}
 0\le r^\infty_{ic}-r_{ic}
 =\lambda_h a_c\,\erfc\!\left(\frac{i-p_c}{2\sigma_h}\right)
 \le\lambda_h a_c\,\erfc\!\left(\frac{s}{4\sigma_h}\right).
 \label{eq:position-residual}
\end{equation}
At $s=32$ and $\sigma_h=4$, $\erfc(2)\simeq0.004678$: the difference is at most
$0.468\%$ of the norm bonus. This is not a relative bound on the total score,
which can include cancellation with the content term. Larger offsets further
reduce the difference.

For a query $i$ and a fixed available set with more than $K$ cells, let $\gamma_\infty$ be
the gap between the $K$th and $(K+1)$st constant-gate scores and
$\delta_{\max}=\max_c\lambda_h a_c\erfc((i-p_c)/(2\sigma_h))$.
If $\gamma_\infty>\delta_{\max}$, replacing the constant gate by $g_h$ leaves the top-$K$ set
unchanged: a selected score falls by at most $\delta_{\max}$, while an
unselected score cannot increase. The raw-token readout is then unchanged too.
Equivalently, routing can change only when $\gamma_\infty\le\delta_{\max}$. It applies to FS's full prefix and
FS-L's retained set at fixed representations.

\paragraph{Fixed-state routing controls.}
On the $16$k native traces described in Appendix~\ref{app:routing-probe},
replacing $g_h$ by one keeps $99.998\%$ of the native top-$K$ cells in the
top-$K$ set in both $340$M models. Removing the entire bonus instead preserves only
$53.9/14.0/14.7\%$ in FS and $73.3/32.5/32.7\%$ in FS-L on S-NIAH-1/2/3.
The norm bonus changes routing substantially, whereas its residual distance
dependence rarely changes the selected set.

\FloatBarrier

\subsection{Cell summaries and reading distance}
\label{app:generalization}\label{app:memory-access}
FS imposes a bottleneck on cell selection: a query selects cells using
mean keys and maximum key norms, then reads the selected cells' original
keys and values. The routing summary need only locate useful evidence;
it need not contain the answer. Sparse readout also bounds how many token
values can interfere, independently of their distance from the query.

\paragraph{A keypoint can mark a cell.}
If a cell contains one nonzero key $\mathbf k_\star$, its mean is
$\mathbf k_\star/s$, while $a_c=\|\mathbf k_\star\|_2$. The content match
is diluted by $1/s$, but the norm bonus preserves the cue's magnitude.
One key can thus make a cell visible to the router. A high-norm distractor
receives the same bonus, so query alignment still decides which cells are read.

\paragraph{Why sparse readout matters after RMSNorm.}
Let $\mathbf y=\sum_{j\in\mathcal S_i}e^{\ell_{ij}}\mathbf v_j$,
$Y=\sum_{j\in\mathcal S_i}e^{\ell_{ij}}$, and $\mathbf g$ be the
readout RMSNorm gain. Then
\begin{equation}
 \mathbf o_i^{\rm lr}
 =\frac{\mathbf g\odot\mathbf y}
 {\sqrt{\|\mathbf y\|_2^2/d+\epsilon Y^2}}.
 \label{eq:normalized-readout}
\end{equation}
When $\|\mathbf y/Y\|_2^2/d\gg\epsilon$ ($\epsilon=10^{-8}$), the scalar
softmax denominator cancels. Normalization therefore removes amplitude dilution,
but irrelevant values can still alter the output direction. Sparse selection restricts
which values enter this sum, rather than merely reducing its softmax
denominator. This distinction matters when relating FreeSolo to analyses of
attention dilution and sparse length generalization~\citep{chiang2022overcoming,vasylenko2026sparse}.

\begin{figure}[t]
\centering
\includegraphics[width=\linewidth,height=0.88\textheight,keepaspectratio]{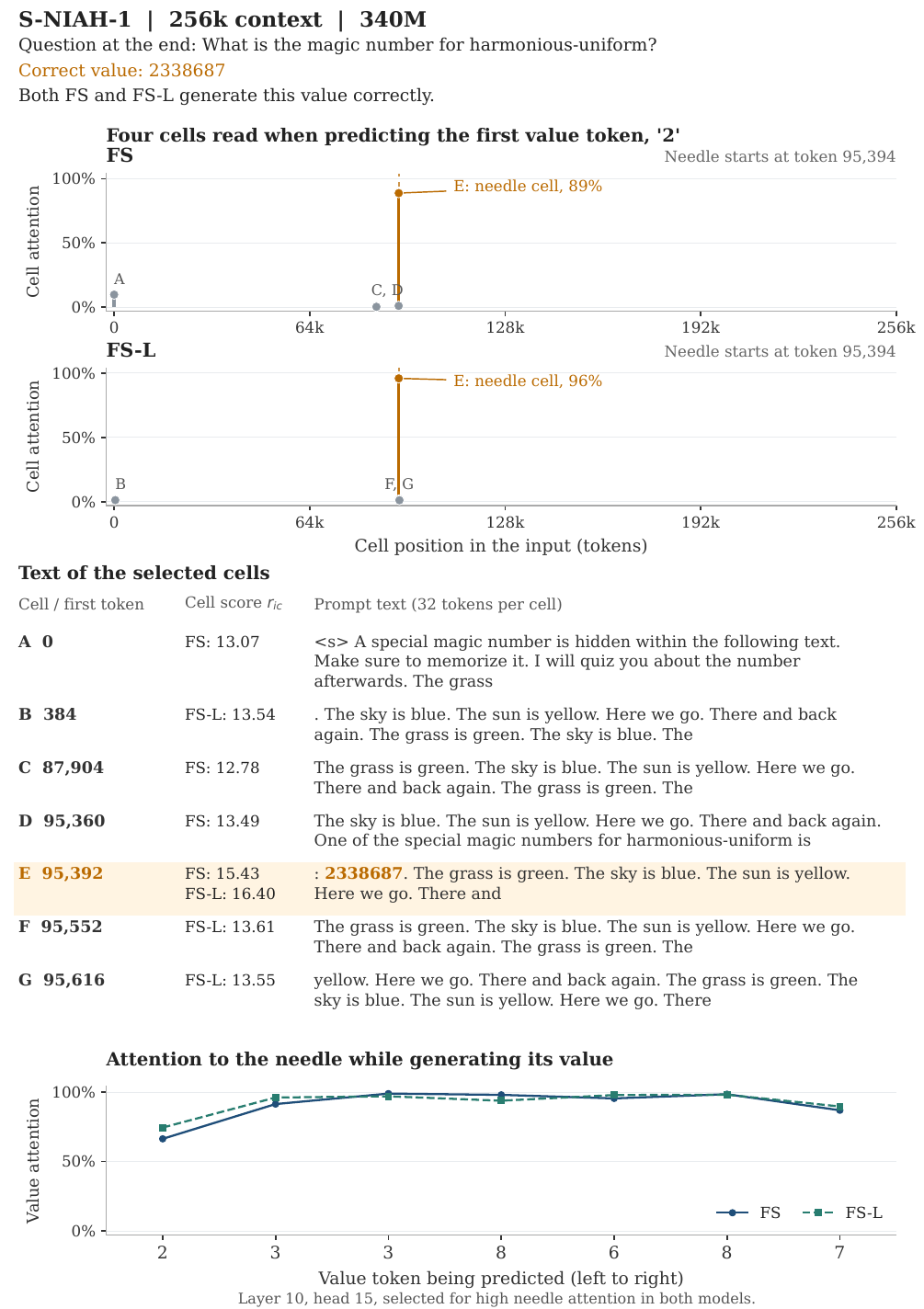}
\caption{\textbf{Retrieving a number at $256$k ($340$M).} Selected cells, routing
scores, prompt text, and attention during generation. Both answers are correct.}
\label{fig:read-distance-340m}
\end{figure}

\paragraph{Reading the needle.}
Figure~\ref{fig:read-distance-340m} traces $340$M FS and FS-L while answering
the same $256$k S-NIAH-1 prompt. Both generate the correct value, $2338687$,
from tokens $95{,}394$--$95{,}400$ in a $262{,}010$-token input.
For a direct comparison, we display the same head in both models: layer $10$,
head $15$, the head that maximizes the smaller of the two models' mean needle
attention during the first answer. These are selected retrieval
examples, not averages across heads or prompts.

The upper plots show the four cells read when predicting the first value
token. A stem's height is the cell's total long-range softmax weight;
orange marks the cell containing the value. The letters link each read to
its decoded $32$-token text, including unrelated cells. Each row gives the
routing score $r_{ic}$ from Eq.~\eqref{eq:qqr-score} for the models that selected
the cell. These scores include the content term and erf prior; they are distinct
from the attention weights plotted above. The lower plot follows attention
to the exact source value as each answer token is predicted, excluding other
tokens in its cell.
In this example, the needle cell receives $89\%/96\%$ of the displayed head's
long-range softmax weight in FS/FS-L at the first value token. Both heads continue to read
the distant value during generation. These weights precede readout RMSNorm
and branch mixing; they do not establish a causal contribution to the answer.

\begin{figure}[t]
\centering
\includegraphics[width=\linewidth]{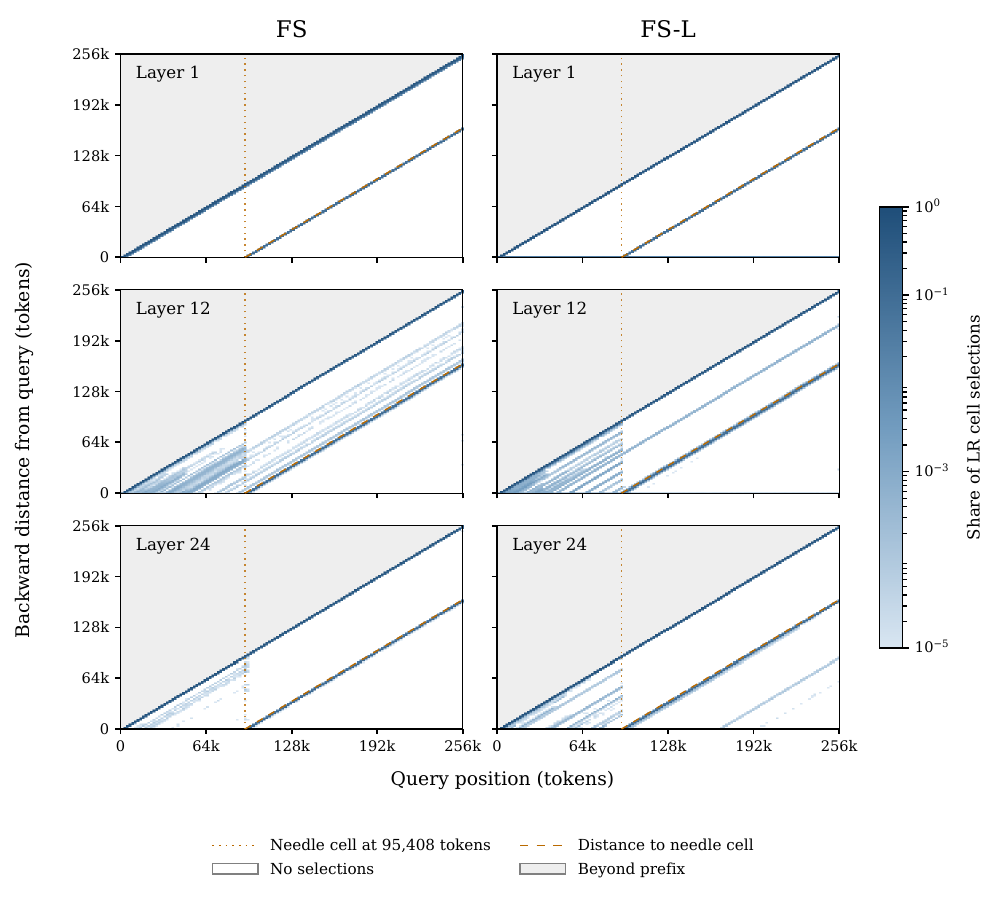}
\caption{\textbf{Long-range read distance at $800$M} on one $256$k S-NIAH-1 prompt.
Orange guides mark the needle position and its distance from each query.}
\label{fig:read-distance-800m}
\end{figure}

\paragraph{Read distance across the prompt.}
Unlike Raven, which measures memory by the tokens written to recurrent
slots~\citep[Definition~4 and Figure~8]{afzal2026raven}, we measure where queries
read. Selecting cell $c$ at query $i$ gives backward distance $i-p_c$ tokens.
Figure~\ref{fig:read-distance-800m} plots this distance against query position
for the $800$M checkpoints on the same $256$k prompt.
Both axes use tokens ($1$k$\;=1024$) and $2048$-token bins. Color shows the
fraction of cell selections per query bin, aggregated over the layer's $16$
heads, on a logarithmic scale. White denotes zero selections; gray lies
beyond the available prefix.

The needle occupies cell $2981$, centered at $p_\star=95{,}408$.
The orange dotted line marks its position; the dashed diagonal follows its
distance $i-p_\star$ from later queries. Diagonal bands indicate repeated
access to fixed earlier cells. Both checkpoints read the needle cell in
layer $1$. At the final prompt query, $166{,}601$ tokens from its center,
layer $24$ selects it in $6/16$ FS heads and $2/16$ FS-L heads.
Over the final $2048$ queries, averaging all layers and heads, $93.6\%$ of FS
selections and $93.1\%$ of FS-L selections lie at least $131{,}072$ tokens
behind the query. These counts show direct access to distant cells, without
measuring their attention weights.

\begin{figure}[t]
\centering
\includegraphics[width=\linewidth,height=0.88\textheight,keepaspectratio]{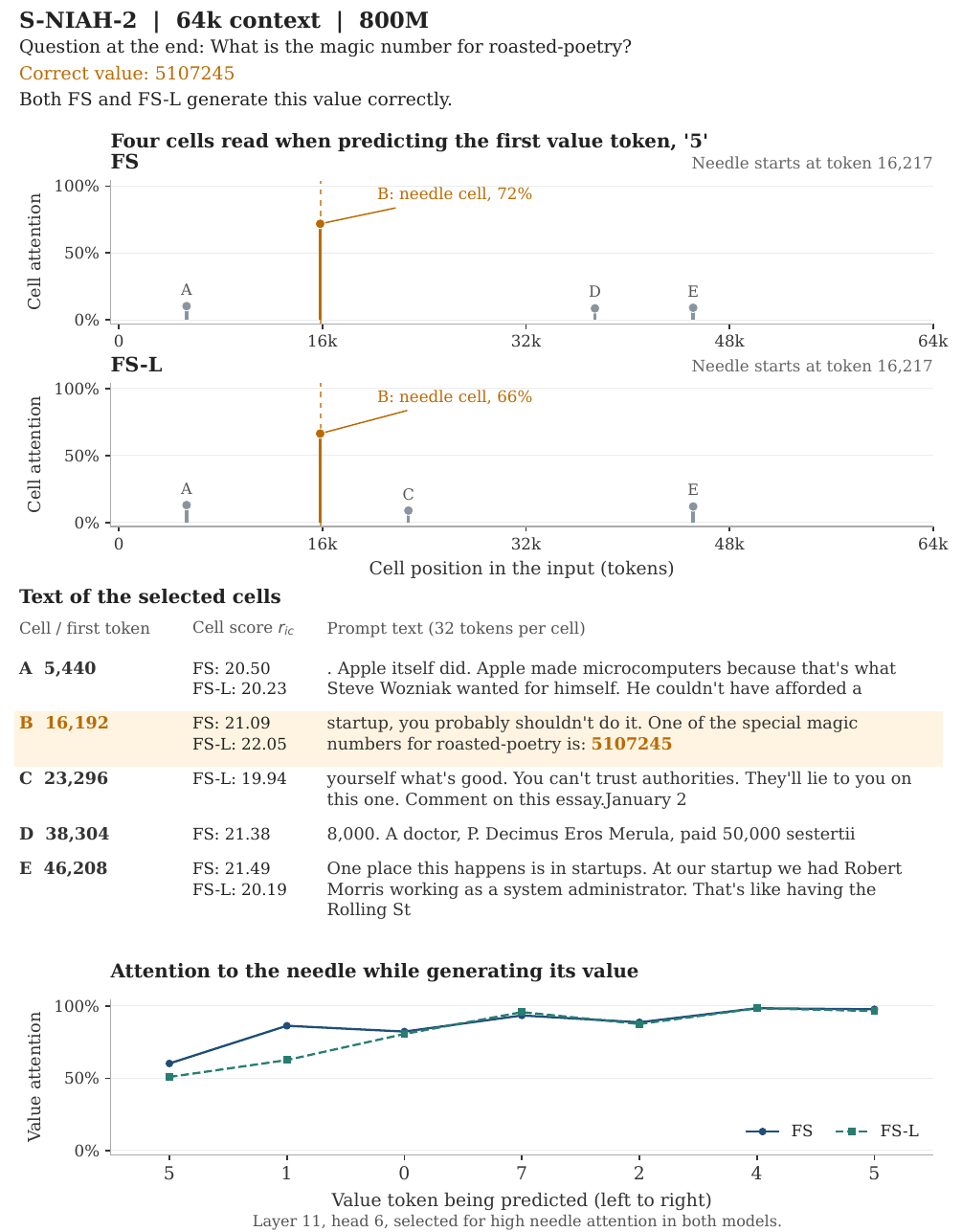}
\caption{\textbf{Retrieving a number at $64$k ($800$M).} Selected cells, routing
scores, prompt text, and attention to the S-NIAH-2 value. Both answers are correct.}
\label{fig:read-distance-s2}
\end{figure}

\begin{figure}[t]
\centering
\includegraphics[width=\linewidth,height=0.88\textheight,keepaspectratio]{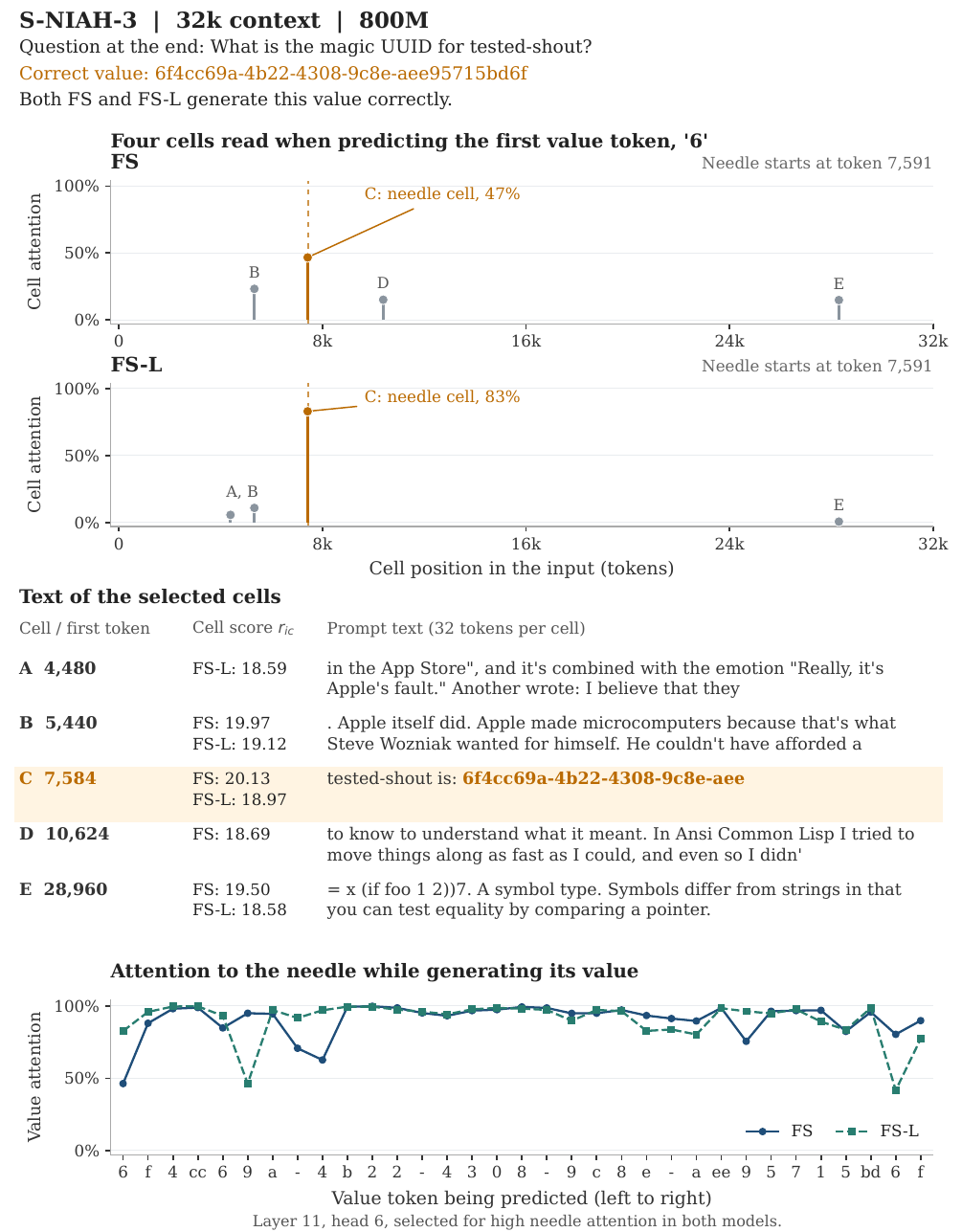}
\caption{\textbf{Retrieving a UUID at $32$k ($800$M).} Selected cells, routing
scores, prompt text, and attention to the S-NIAH-3 value. Both answers are correct.}
\label{fig:read-distance-s3}
\end{figure}

\paragraph{Retrieval on harder tasks.}
Figures~\ref{fig:read-distance-s2} and~\ref{fig:read-distance-s3} show selected
successful S-NIAH-2/3 examples at $64$k/$32$k with the same $800$M checkpoints.
Each pair uses the same prompt and needle position; both models generate
the correct value. The common-head rule above selects layer $11$, head $6$
for both tasks. For S-NIAH-2, the needle cell near token $16{,}217$ receives
$72\%/66\%$ of this head's long-range softmax weight in FS/FS-L at the first value token.
For S-NIAH-3, the cell near token $7{,}591$ receives $47\%/83\%$.
The displayed text shows the number or UUID prefix alongside the selected
essay excerpts. The UUID spans two cells; only its first cell is read at the
snapshot shown, and the next cell is selected later during generation.
The lower plots follow attention to the full source value across the answer.

\FloatBarrier

\subsection{Norm pruning and retrieval coverage}
\label{app:routing-probe}
We measure how much norm pruning reduces the available prefix and how it
affects retrieval. Key norm is a learned pruning heuristic: a smaller-norm cell
can still be relevant to a future query. We therefore evaluate source access,
selection coverage, and answer accuracy separately.

\paragraph{Separate retention from selection.}
A non-recent cell survives pruning when fewer than $P$ cells outrank it by
$a_c$. It is then read only if fewer than $K$ retained cells outrank it by
$r_{ic}$. Raising $K$ cannot recover a cell already pruned. At fixed scores,
retaining every FS top-$K$ cell preserves its selected-token readout, but
partial overlap need not preserve the answer.

\paragraph{Probe protocol.}
We trace the FS and FS-L checkpoints of Table~\ref{tab:ruler} at both scales on $50$
examples per S-NIAH task at $2$k/$4$k/$8$k/$16$k, totaling $600$ prompts per model.
Each trace covers all $24$ layers and $16$ heads on the final $16$ prompt
positions. The forward pass uses each model's native routing. Alternative scores are computed on those same representations without changing the forward
pass. Source-access measurements exclude prompts whose answer source overlaps
the local window of any observed query. We average head/query rates within each
prompt, then average prompts; heads are not independent observations.

\paragraph{Source availability and selection.}
Table~\ref{tab:source-access} distinguishes source retention from query-dependent
selection. FS keeps the full prefix, so its retention column applies
the FS-L rule offline to FS keys. Its selection column uses native FS routing.
FS-L's columns both describe its native operation. These probes and the
eligible-prompt answer counts below use the full-history masked implementation
of FS-L, rather than the bounded cache used for the main accuracy tables.

\begin{table}[ht]
\caption{\textbf{Source-cell availability and selection at $16$k.} The source is
the cell containing the answer value. Rates average over all heads and the final
$16$ prompt positions within each prompt, then over prompts; FS (offline) applies
the FS-L retention rule to FS keys.}
\label{tab:source-access}
\begin{center}
\small\setlength{\tabcolsep}{9pt}
\begin{tabular}{@{}lr r r r r@{}}
\toprule
& & \multicolumn{2}{c}{Source retained (\%)} & \multicolumn{2}{c}{Source selected (\%)} \\
\cmidrule(lr){3-4}\cmidrule(lr){5-6}
Task & Prompts & FS (offline) & FS-L & FS & FS-L \\
\midrule
\multicolumn{6}{@{}l}{\emph{$340$M scale}} \\
S-NIAH-1 & $47$ & $22.31$ & $27.98$ & $38.15$ & $17.55$ \\
S-NIAH-2 & $50$ & $18.34$ & $19.87$ & $1.27$ & $1.30$ \\
S-NIAH-3 & $47$ & $24.64$ & $25.76$ & $2.49$ & $2.65$ \\
\midrule
\multicolumn{6}{@{}l}{\emph{$800$M scale}} \\
S-NIAH-1 & $47$ & $23.07$ & $22.03$ & $29.21$ & $14.99$ \\
S-NIAH-2 & $50$ & $16.22$ & $17.03$ & $2.00$ & $2.08$ \\
S-NIAH-3 & $47$ & $26.29$ & $26.42$ & $3.55$ & $3.52$ \\
\bottomrule
\end{tabular}
\end{center}
\end{table}

At $340$M, under the same norm rule, FS-L improves source availability by
$5.66/1.54/1.11$ percentage points on S-NIAH-1/2/3 (computed before rounding).
Paired-prompt bootstrap
$95\%$ intervals are $[5.05,6.31]$, $[1.14,1.93]$, and $[0.72,1.49]$
($5000$ resamples). These compare the two checkpoints and do not separate
pruning adaptation from additional training. Native source selection
is substantially lower in FS-L on S-NIAH-1 and similar on the harder tasks.
Among head/query pairs where the source survives, FS-L selects it in
$63.8/6.6/10.3\%$ of cases, averaging the conditional rate within each prompt.
Source access does not guarantee a correct answer: on the same eligible prompts,
greedy answers are correct on $47/47$, $9/50$, and $0/47$ for FS and $40/47$,
$9/50$, and $0/47$ for FS-L on S-NIAH-1/2/3 at $16$k.

At $800$M, the corresponding source-availability changes are
$-1.04/0.82/0.13$ percentage points, with paired $95\%$ intervals
$[-2.06,0.07]$, $[0.47,1.16]$, and $[-0.35,0.64]$.
The increase at $340$M therefore does not generalize uniformly across scales.
Native S-NIAH-1 source selection is again lower in FS-L, while S-NIAH-2/3
selection changes little. FS-L selects a surviving source in
$69.3/12.3/13.4\%$ of eligible head/query pairs, averaging within each prompt.
On S-NIAH-2/3, FS selects the source in only $1.3$--$3.6\%$ of head/query pairs
and FS-L selects a retained source in $6.6$--$13.4\%$, so selection rather than
retention limits the harder tasks.

\paragraph{Compression.}
For each head, define the retained fraction
$\eta_i=|\mathcal A_i|/|\mathcal C_i|$. At $4$k/$8$k/$16$k, native FS-L retains
$58.8/29.0/14.3\%$ of eligible cells across these probes, corresponding to
$1.7/3.4/7.0\times$ reductions in cells available for scoring. All lengths use
$P=64$ and $W=8$. These fractions concern cell counts; Appendix~\ref{app:bounded-cache}
measures physical storage.

\FloatBarrier

\subsection{Continued pretraining}
\label{app:sparse-fail}\label{app:hard-selection}
CPT recovers much of the S-NIAH-1 accuracy lost under immediate pruning.
Harder-task changes also occur with unpruned continuation and vary across
seeds. These results support adapting an FS checkpoint to the pruning
constraint through short CPT.

\paragraph{Learning with hard indices.}
As in Section~\ref{sec:longrange}, top-$P$ and top-$K$ indices receive no loss
gradient, and we use no auxiliary retention loss or straight-through estimator.
A key receives no gradient through an access that excludes it, and the remaining
gradients need not increase its norm; scores and norm ranks change only through
the shared projections and contextual states.

CPT starts from FS's retrieval representations and keeps pruning in the loop
at the same $4096$-token context. The constraint already binds during training:
the last query block has $124$ eligible cells, exceeding $P+W=72$.
Optimization settings are in Appendix~\ref{app:hyperparams}.

\paragraph{Continuation controls.}
Table~\ref{tab:continuation-340m} compares pruning and continuation at
$340$M using the same first $50$ examples per task at $4$k. Immediate pruning
applies FS-L's $P=64$ retention rule without adaptation. The seeded controls
start from the same FS checkpoint with a fresh optimizer, the same corpus
and schedule, and $1500$ steps ($0.786$B tokens); FS uses full-prefix routing
and FS-L keeps pruning in training and evaluation. The main FS-L checkpoint
uses $1400$ steps. We repeat seed $42$ with the same data order and run seed
$43$ once; changing the seed also changes data sampling.

\begin{table}[ht]
\caption{\textbf{Pruning recovery and seed variation at $340$M.}
Accuracy (\%) on $50$ shared prompts per task at $4$k, the training context.
FS-L is the $1400$-step checkpoint of Table~\ref{tab:ruler}; seeded rows use
$1500$ CPT steps without (FS) or with (FS-L) pruning.}
\label{tab:continuation-340m}
\centering\small\setlength{\tabcolsep}{6pt}
\begin{tabular}{@{}lrrr@{}}
\toprule
Model & S-NIAH-1 & S-NIAH-2 & S-NIAH-3 \\
\midrule
FS base & $100$ & $74$ & $14$ \\
Immediate pruning & $62$ & $72$ & $14$ \\
FS-L & $90$ & $64$ & $14$ \\
\midrule
FS, seed 42 (run 1) & $100$ & $50$ & $16$ \\
FS, seed 42 (run 2) & $100$ & $52$ & $20$ \\
FS, seed 43 & $100$ & $74$ & $6$ \\
\midrule
FS-L, seed 42 (run 1) & $82$ & $44$ & $12$ \\
FS-L, seed 42 (run 2) & $76$ & $46$ & $18$ \\
FS-L, seed 43 & $82$ & $82$ & $8$ \\
\bottomrule
\end{tabular}
\end{table}

At $4$k, immediate pruning reduces S-NIAH-1 from $100$ to $62$. FS-L
recovers $90$, while the seed-$42$ adapted runs recover $20$ and $14$
percentage points (paired-prompt bootstrap $95\%$ intervals $[4,36]$ and
$[0,28]$). All three unpruned continuations retain $100$. S-NIAH-2 declines
under seed $42$ even without pruning, whereas seed $43$ reaches $82$ with
pruning versus $74$ without it. Thus CPT supports S-NIAH-1 recovery, but
harder-task changes vary across seeds and also occur without pruning.
Intervals use $5000$ prompt resamples and do not estimate seed uncertainty.

Harder-task gains also occur without pruning at $800$M: after $3000$ CPT
steps ($1.57$B tokens), FS's $16$k S-NIAH-2/3 accuracy rises from $72/32$
to $84/38$ on $50$ shared prompts per task.

\paragraph{Budget-matched continuation of unchanged architectures.}
To assess the effect of additional training alone,
Table~\ref{tab:matched-cpt} reports selected $800$M-scale baselines and FS
continued from their Table~\ref{tab:ruler} checkpoints with plain next-token
loss. These controls match FS-L's $3000$ updates of $524{,}288$ tokens each
($1.57$B total), using FineWeb-Edu at $4096$-token context
(Appendix~\ref{app:hyperparams}). The table also includes a separate
single-forest width-$32$ ablation of FS-B's recall-weighted recipe:
$1600$ updates of $262{,}144$ tokens each ($0.42$B total, $27\%$ of the
matched budget), at the same context length (Appendix~\ref{app:fsb-cpt}).
Each before/after pair uses identical prompts and task sets: means cover
$12$ tasks at $2$k/$4$k and $13$ at $8$k/$16$k, following the protocol of
Table~\ref{tab:ruler}.

Plain CPT does not improve the baselines or unpruned FS consistently across
all evaluated lengths. Only FS-B's recall-weighted CPT improves the RULER
mean at all four lengths in Table~\ref{tab:matched-cpt}, as measured by the
single-forest width-$32$ ablation: $40.9/36.0/30.4/28.2$ becomes
$41.3/37.2/31.6/30.4$. FS-L recovers much of the S-NIAH-1 accuracy lost
under immediate pruning (Table~\ref{tab:continuation-340m}), but its suite
mean remains below base FS at every length in Table~\ref{tab:matched-cpt}.

\begin{table}[ht]
\caption{\textbf{Only FS-B's weighted CPT improves the RULER mean at all four lengths.}
Suite-mean changes at $800$M scale
($+$\,CPT minus base, points, on the same prompts and protocols as
Table~\ref{tab:ruler}; the base means are the Table~\ref{tab:ruler} rows).
Each ``$+$CPT'' arm continues the Table~\ref{tab:ruler} checkpoint of the
same architecture with plain next-token loss at the FS-L budget ($1.57$B
tokens). The FS-L arm instead keeps the
pruning loop in training at the same budget. The single-forest ablation uses
FS-B's recall-weighted CPT, spending $0.42$B tokens ($27\%$;
Appendix~\ref{app:fsb-cpt}), and compares the full-prefix width-$32$ read
before and after CPT (base means $40.9/36.0/30.4/28.2$).
FS-B itself denotes the two-forest design of Table~\ref{tab:ruler}.}
\label{tab:matched-cpt}
\centering\footnotesize\setlength{\tabcolsep}{4pt}
\begin{tabular}{@{}l*{4}{r}@{}}
\toprule
Continuation arm & $\Delta\,2$k & $\Delta\,4$k & $\Delta\,8$k & $\Delta\,16$k \\
\midrule
SWA, $743$M, plain & $-1.4$ & $+0.4$ & $+0.2$ & $-0.2$ \\
Raven-full, $723$M, plain & $-7.2$ & $-5.6$ & $-1.4$ & $+0.8$ \\
Raven, $726$M, plain & $-2.2$ & $-1.0$ & $-2.8$ & $-1.9$ \\
NSA, $1128$M, plain & $-1.2$ & $-2.2$ & $+0.3$ & $+0.1$ \\
\midrule
FS, $743$M, plain & $-1.6$ & $-1.3$ & $-1.0$ & $0.0$ \\
FS, $743$M $\to$ FS-L (pruning) & $-2.2$ & $-2.0$ & $-2.0$ & $-0.8$ \\
Single-forest W32 ablation, $743$M, weighted & $+0.4$ & $+1.2$ & $+1.2$ & $+2.2$ \\
\bottomrule
\end{tabular}
\end{table}

\paragraph{Training from scratch.}
CPT gives a higher RULER mean than training from scratch
at all four lengths (Table~\ref{tab:cpt-vs-scratch-340m}). The gain comes from
S-NIAH-2/3, with a tradeoff on S-NIAH-1 and downstream accuracy. At $16$k, the mean rises from $9.6$ to $10.7$, while S-NIAH-1 falls
from $93$ to $87$. The eight-task downstream average falls from $52.2$ to
$51.0$.

At $800$M, training with pruning from scratch reaches a downstream average of
$56.8$ but S-NIAH-1 scores of only $32$, $16$, $8$, and $0$ at $2$k, $8$k,
$16$k, and $32$k ($25$ examples), whereas CPT retrieves well
(Table~\ref{tab:ruler}); we therefore use CPT at both scales.

\begin{table}[h]
\caption{\textbf{FS-L training recipes at $340$M} ($0$--$100$).
Both use pruning during evaluation, with $100$ examples per RULER task and length.
Bold marks the better recipe.}
\label{tab:cpt-vs-scratch-340m}
\begin{center}
\footnotesize\setlength{\tabcolsep}{3pt}
\resizebox{\textwidth}{!}{\begin{tabular}{@{}l cccc cccc cccc cccc c@{}}
\toprule
& \multicolumn{4}{c}{S-NIAH-1} & \multicolumn{4}{c}{S-NIAH-2} & \multicolumn{4}{c}{S-NIAH-3} & \multicolumn{4}{c}{RULER mean} & Down. \\
\cmidrule(lr){2-5}\cmidrule(lr){6-9}\cmidrule(lr){10-13}\cmidrule(lr){14-17}
Recipe & $2$k & $4$k & $8$k & $16$k & $2$k & $4$k & $8$k & $16$k & $2$k & $4$k & $8$k & $16$k & $2$k & $4$k & $8$k & $16$k & avg \\
\midrule
From scratch & $\mathbf{100}$ & $\mathbf{97}$ & $\mathbf{98}$ & $\mathbf{93}$ & $78$ & $49$ & $8$ & $3$ & $53$ & $11$ & $3$ & $0$ & $27.8$ & $19.2$ & $11.7$ & $9.6$ & $\mathbf{52.2}$ \\
CPT & $99$ & $88$ & $90$ & $87$ & $\mathbf{99}$ & $\mathbf{70}$ & $\mathbf{34}$ & $\mathbf{20}$ & $\mathbf{70}$ & $\mathbf{17}$ & $\mathbf{9}$ & $\mathbf{1}$ & $\mathbf{32.4}$ & $\mathbf{21.6}$ & $\mathbf{15.5}$ & $\mathbf{10.7}$ & $51.0$ \\
\bottomrule
\end{tabular}}
\end{center}
\end{table}

\FloatBarrier

\subsection{Retention budgets and multi-fact retrieval}
\label{app:learned-retention}\label{app:compression-rate}
FS-L must retain information before knowing which question will be asked.
The adapted budget sweep measures the resulting accuracy--memory tradeoff.
The multiple-question test checks whether the models can answer questions
about several facts near the training length.

\paragraph{The question distribution sets the budget.}
A small retained set can cover likely future questions when useful information
is concentrated in a few cells. Consider instead $F$ independent key--value
facts in separate cells, with a later question choosing one uniformly.
Keeping $m\le F$ source cells gives direct access to only $m/F$ of the possible
targets. Contextual representations can copy information between cells, so
this ratio is not an answer-accuracy bound. The accessible state includes
all heads and layers, local buffers, and their contextual representations.

Compression does not remove this limit. If the $F$ values are independent and
uniform over $V$ outcomes, the question reveals only which value to retrieve,
and the state holds $b$ bits, Fano's inequality bounds the mean accuracy by
$\bar p\le(1+b/F)/\log_2V$; any fixed accuracy with $\bar p\log_2V>1$ therefore
needs $b=\Omega(F)$.
This motivates testing memory budgets and preserving more cells when information
is dense. It does not identify the bottleneck in a trained model: a wrong answer
can reflect lost information, failed routing, or failure to use available evidence.

\paragraph{Adapted retention budgets.}
From the same base FS checkpoint, we adapt with $P=16$ and $P=32$, keeping
$W=8$, $K=4$, and the same $1500$-step schedule. Each model evaluates with its training budget on
the same $50$ prompts (Table~\ref{tab:adapted-budgets}). At $2$k, S-NIAH-1
falls from $100$ at $P=64$ to $98$ at $P=32$ and $84$ at $P=16$, but accuracy
is not monotone in $P$: at $16$k, $P=64/32/16$ score $74$--$80/66/68$, using the
two seed-$42$ controls at $P=64$. S-NIAH-2/3 do not improve consistently with $P$.
These independently adapted operating points measure a training--memory
tradeoff, not the effect of changing the budget in one fixed model.
Only $P=64$ has repeated continuations. The local and temporary buffers also
contribute to physical state, so changing $P$ by a factor of four does not
change total memory by that factor.

\begin{table}[ht]
\caption{\textbf{Adapted retention budgets at $340$M.} Each model trains and evaluates with its stated $P$.}
\label{tab:adapted-budgets}
\centering\footnotesize\setlength{\tabcolsep}{3pt}
\begin{tabular}{@{}l*{12}{r}@{}}
\toprule
& \multicolumn{4}{c}{S-NIAH-1} & \multicolumn{4}{c}{S-NIAH-2} & \multicolumn{4}{c}{S-NIAH-3} \\
\cmidrule(lr){2-5}\cmidrule(lr){6-9}\cmidrule(lr){10-13}
Model & $2$k & $4$k & $8$k & $16$k & $2$k & $4$k & $8$k & $16$k & $2$k & $4$k & $8$k & $16$k \\
\midrule
$P=16$ & $84$ & $64$ & $68$ & $68$ & $\mathbf{100}$ & $\mathbf{56}$ & $\underline{24}$ & $\mathbf{14}$ & $56$ & $14$ & $4$ & $\mathbf{2}$ \\
$P=32$ & $\underline{98}$ & $\underline{78}$ & $\underline{72}$ & $66$ & $96$ & $\underline{50}$ & $\mathbf{32}$ & $\underline{10}$ & $\mathbf{74}$ & $\mathbf{20}$ & $\underline{6}$ & $\mathbf{2}$ \\
$P=64$, run 1 & $\mathbf{100}$ & $\mathbf{82}$ & $\mathbf{82}$ & $\mathbf{80}$ & $\underline{98}$ & $44$ & $22$ & $2$ & $\underline{66}$ & $12$ & $\mathbf{8}$ & $\mathbf{2}$ \\
$P=64$, run 2 & $\mathbf{100}$ & $76$ & $70$ & $\underline{74}$ & $96$ & $46$ & $20$ & $2$ & $58$ & $\underline{18}$ & $\mathbf{8}$ & $\mathbf{2}$ \\
\bottomrule
\end{tabular}
\end{table}

\begin{figure}[ht]
\centering
\includegraphics[width=\linewidth]{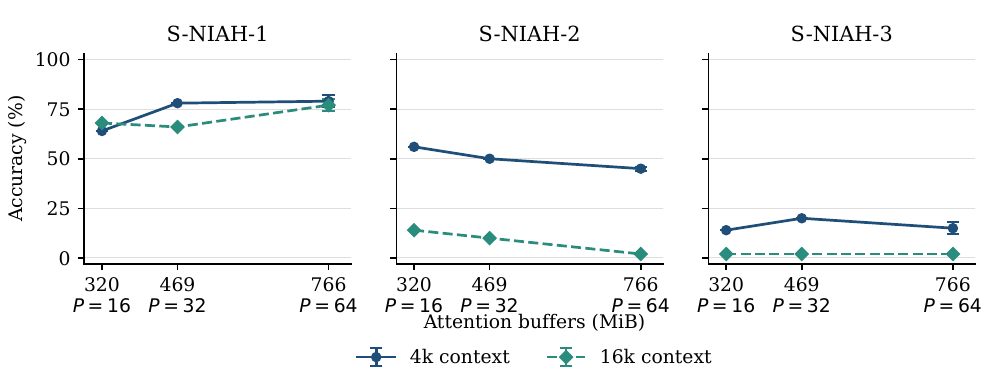}
\caption{\textbf{Accuracy versus attention storage at $340$M.}
The $P=64$ points show the mean and range of two seed-$42$ continuations.}
\label{fig:retention-tradeoff}
\end{figure}

Figure~\ref{fig:retention-tradeoff} relates these scores to the allocated
attention buffers of the streaming cache (Table~\ref{tab:budget-resources},
protocol of Appendix~\ref{app:bounded-cache}): reducing $P$ fourfold gives a
$2.4\times$ reduction in total buffers (which do not depend on length), while
decode stays at $88$--$89$ tokens/s. The $P=64$ row of Table~\ref{tab:budget-resources}
uses the Table~\ref{tab:ruler} checkpoint, while the $P=64$ accuracies in
Figure~\ref{fig:retention-tradeoff} average the two seed-$42$ continuations.

\begin{table}[ht]
\caption{\textbf{Retention budget and bounded inference at $340$M, $256$k.}}
\label{tab:budget-resources}
\centering\small\setlength{\tabcolsep}{7pt}
\begin{tabular}{@{}rrrrr@{}}
\toprule
$P$ & Buffers (MiB) & Peak (GiB) & Prefill (Ktok/s) & Decode (tok/s) \\
\midrule
$16$ & $\mathbf{320.0}$ & $\mathbf{1.34}$ & $2.13$ & $\underline{88.4}$ \\
$32$ & $\underline{468.8}$ & $\underline{1.49}$ & $\mathbf{2.19}$ & $\mathbf{88.8}$ \\
$64$ & $766.2$ & $1.77$ & $\underline{2.17}$ & $88.3$ \\
\bottomrule
\end{tabular}
\end{table}

\paragraph{Several facts at fixed context.}
\label{app:multiple-questions}
Each synthetic prefix holds $F$ distinct key--value facts with seven-digit
values, each outside the final $528$ prefix tokens. The question is revealed
only after the prefix, and each question continues an independent copy of the
prefix state. We generate $16$ tokens greedily and score whether the target
value appears, with $50$ prefixes per condition, querying every fact for
$F\le8$ and eight sampled facts for $F=16$. Prefixes contain $4253$--$4277$
tokens at $340$M and $3946$--$3971$ at $800$M, where prefix, question and
answer fit within $4096$ tokens.

\begin{table}[ht]
\caption{\textbf{Multiple-question retrieval.} Accuracy (\%) on
shared prefixes: an answer is correct if the target value appears in $16$ greedy
tokens; $50$ prefixes per condition. Bold and underline mark the best and
second-best score within each scale, including ties. FS-B prefills each
question afresh; at $340$M this raises FS and FS-L by about $2$ points
relative to the shared-prefix rows.}
\label{tab:multiple-questions}
\begin{center}
\small\setlength{\tabcolsep}{10pt}
\begin{tabular}{@{}lrrrrr@{}}
\toprule
Facts per prefix & $1$ & $2$ & $4$ & $8$ & $16$ \\
Questions & $50$ & $100$ & $200$ & $400$ & $400$ \\
\midrule
\multicolumn{6}{@{}l}{\emph{$340$M, $4253$--$4277$ prefix tokens}} \\
FS & $\underline{94}$ & $41$ & $19$ & $7.75$ & $2.75$ \\
FS-L & $90$ & $\underline{48}$ & $\underline{21}$ & $\underline{8.00}$ & $\underline{3.25}$ \\
FS-B & $\mathbf{100}$ & $\mathbf{56}$ & $\mathbf{28.5}$ & $\mathbf{13.0}$ & $\mathbf{4.25}$ \\
\midrule
\multicolumn{6}{@{}l}{\emph{$800$M, strictly within $4096$ tokens}} \\
FS & $\mathbf{100}$ & $49$ & $26$ & $11.5$ & $7.25$ \\
FS-L & $\underline{96}$ & $53$ & $21$ & $10$ & $4$ \\
TF++ & $88$ & $56$ & $\mathbf{38}$ & $\mathbf{22}$ & $\mathbf{19.75}$ \\
Raven & $\mathbf{100}$ & $\underline{57}$ & $28.5$ & $9$ & $2$ \\
FS-B & $\mathbf{100}$ & $52$ & $32$ & $16.5$ & $8.75$ \\
Raven-full & $\mathbf{100}$ & $\mathbf{66}$ & $\underline{33.5}$ & $\underline{21.75}$ & $\underline{13}$ \\
\bottomrule
\end{tabular}
\end{center}
\end{table}

FS, FS-L, and FS-B pass a $90\%$ single-fact threshold at both scales, but no
model passes it with two or more facts (Table~\ref{tab:multiple-questions});
FS-B scores highest at $340$M; at $800$M with four or more facts,
full-attention TF++ and Raven-full lead instead. The
decline also occurs with unpruned access, with full-attention TF++ and with
Raven and Raven-full, and at $800$M entirely within the training context, so it is not a
pruning or length-extrapolation limit alone. At $800$M with two facts, most
wrong answers contain another value from the same prefix ($42/47$ for FS-L,
$46/51$ for FS, $48/48$ for FS-B, $36/43$ for Raven and $36/44$ for TF++), consistent with
confusion among stored facts. Because unpruned FS and full-attention TF++
decline in the same way, this test measures interference among facts near the
training length rather than the retention limit above.

\FloatBarrier

\clearpage
\FloatBarrier
\section{FS-B details}
\label{app:fsb}
Section~\ref{sec:fsb} defines FS-B. This appendix explains how the forests are
built and searched, what they cost, and the CPT recipe, and relates FS-B to
chunk and hierarchical retrieval.

\subsection{Streaming forests}
\label{app:fsb-method}
\paragraph{Causal availability.}
Cells contain $s=32$ tokens and query blocks contain $B=128$ tokens. At each
block boundary, cells completed before that boundary are admitted to the
forests by Eq.~\eqref{eq:fsb-admission}. The forests and their summaries then
stay fixed throughout the block, while each token searches independently.
Prefill and cached decoding use the same block-boundary availability as FS
training. Filtering a cell from the second forest does not remove it from the first,
nor its tokens from the local window. The threshold $\tau_{\ell h}$ is fixed
per checkpoint, so the second forest need not hold half of a particular
prefix.

\paragraph{Chunks.}
Fix one head and one forest at a block boundary, and let $c_1,\dots,c_M$ be
its admitted cells in arrival order, with mean keys $\bar{\vb k}_c$, key norms
$a_c$ and positions $p_c$ (Eq.~\eqref{eq:cells}). Write the binary
decomposition $M=\sum_{t=1}^{r}2^{k_t}$ with $k_1>\dots>k_r$. The active chunks
are the consecutive runs
\begin{equation}
  \label{eq:fsb-chunks}
  \mathcal G_t=\{c_j: o_t<j\le o_t+2^{k_t}\},\qquad o_t=\textstyle\sum_{u<t}2^{k_u},
\end{equation}
so older cells sit in larger chunks; $13=8+4+1$ gives $\{c_1,\dots,c_8\}$,
$\{c_9,\dots,c_{12}\}$ and $\{c_{13}\}$. A chunk never changes once complete,
so its tree is built once and reused until two equal chunks merge into the
next larger one.

\paragraph{Splitting a chunk.}
Each node $v$ holds a cell set $\mathcal C(v)$ of size $n_v$; the root of
$\mathcal G_t$ holds all of $\mathcal G_t$. A node with $n_v>1$ cells centers
their mean keys, $\vb x_c=\bar{\vb k}_c-\boldsymbol\mu_v$, and estimates a principal
direction with two power iterations started at the farthest cell:
\begin{equation}
  \label{eq:fsb-split}
  \vb u^{(0)}\propto\vb x_{c^\star},\quad c^\star=\argmax_{c\in\mathcal C(v)}\norm{\vb x_c}_2,
  \qquad
  \vb u^{(t+1)}\propto\sum_{c\in\mathcal C(v)}\big(\vb x_c^\top\vb u^{(t)}\big)\,\vb x_c,
  \quad t=0,1.
\end{equation}
Cells are sorted by $\vb x_c^\top\vb u^{(2)}$; the first $\lceil n_v/2\rceil$
form the left child and the rest the right child. Recursion stops at single
cells, so a chunk of $2^{k}$ cells gives a perfect binary tree of height $k$
whose leaves are ordered by content rather than position.

\paragraph{Node summaries.}
Every chunk-tree node stores
\begin{equation}
  \label{eq:fsb-node}
  \boldsymbol\mu_v=\frac{1}{n_v}\sum_{c\in\mathcal C(v)}\bar{\vb k}_c,\quad
  \rho_v=\max_{c\in\mathcal C(v)}\norm{\bar{\vb k}_c-\boldsymbol\mu_v}_2,\quad
  A_v=\max_{c\in\mathcal C(v)}a_c,\quad
  p_v^{\min}=\min_{c\in\mathcal C(v)}p_c,
\end{equation}
the quantities scored by Eq.~\eqref{eq:fsb-bound}.

\paragraph{Joining chunk roots.}
A chain joins the chunk roots from the smallest chunk upward: $J_r$ is the
root of $\mathcal G_r$, and for $t=r-1,\dots,1$ the join node $J_t$ has children
$\operatorname{root}(\mathcal G_t)$ and $J_{t+1}$; $J_1$ is the forest root. A join
node $v$ with children $a,b$ stores
\begin{equation}
  \label{eq:fsb-join}
  n_v=n_a+n_b,\quad
  \boldsymbol\mu_v=\frac{n_a\boldsymbol\mu_a+n_b\boldsymbol\mu_b}{n_v},\quad
  \rho_v=\max_{u\in\{a,b\}}\big(\rho_u+\norm{\boldsymbol\mu_u-\boldsymbol\mu_v}_2\big),
\end{equation}
with $A_v=\max(A_a,A_b)$ and $p_v^{\min}=\min(p_a^{\min},p_b^{\min})$. By the
triangle inequality its ball encloses every cell mean below it, so
Eq.~\eqref{eq:fsb-bound} remains an upper bound at join nodes. The root of
$\mathcal G_t$ sits at depth $\min(t,r-1)$ and its tree has height
$k_t\le k_1-t+1$, so the forest height is $k_1+[r\ge2]=\lceil\log_2 M\rceil$.

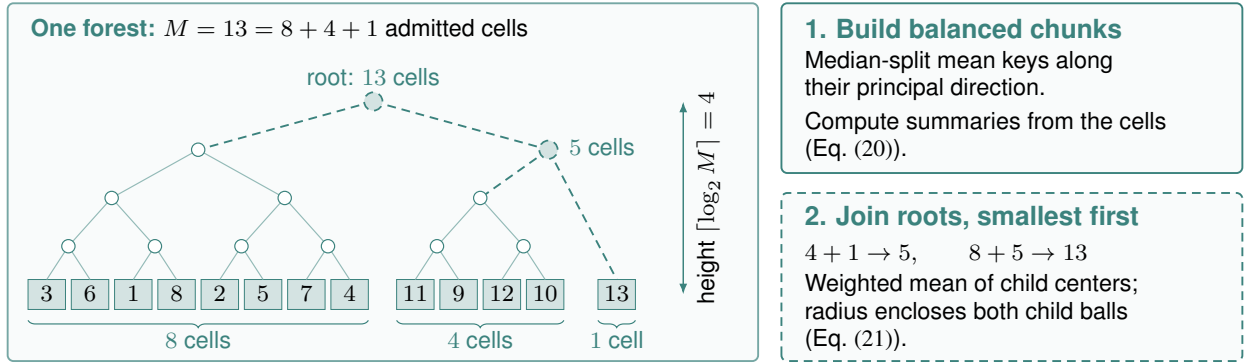
\begin{figure}[!htb]
\centering
\definecolor{mypool}{HTML}{3B827D}
\resizebox{\linewidth}{!}{\begin{tikzpicture}[x=1cm,y=1cm,font=\sffamily\fontsize{8}{9.4}\selectfont,
  panel/.style={rounded corners=2pt,line width=.65pt},
  title/.style={font=\sffamily\bfseries\fontsize{9}{10}\selectfont,anchor=west},
  small/.style={font=\sffamily\fontsize{8}{9.4}\selectfont},
  tnode/.style={circle,draw=mypool,fill=white,line width=.4pt,minimum size=.16cm,inner sep=0pt},
  jnode/.style={tnode,densely dashed,fill=mypool!22,line width=.6pt,minimum size=.22cm},
  leaf/.style={draw=mypool,fill=mypool!22,line width=.3pt,minimum width=.45cm,
    minimum height=.36cm,inner sep=0pt,text height=1.2ex,text depth=.3ex},
  edge/.style={mypool!55,line width=.4pt},
  jedge/.style={draw=mypool,line width=.65pt,densely dashed},
  every node/.style={inner sep=1.5pt}
]
\path[use as bounding box] (0,-.08) rectangle (15,4.40);

\draw[panel,draw=mypool!70,fill=mypool!3] (.08,.02) rectangle (9.10,4.32);
\node[small,anchor=west] at (.30,4.02)
  {\textcolor{mypool}{\bfseries One forest:} $M=13=8+4+1$ admitted cells};

\foreach \k/\a in {0/3,1/6,2/1,3/8,4/2,5/5,6/7,7/4}{
  \coordinate (A\k) at ({.55+.52*\k},.82);
  \node[leaf] at (A\k) {$\a$};
}
\foreach \k/\a in {0/11,1/9,2/12,3/10}{
  \coordinate (B\k) at ({4.98+.52*\k},.82);
  \node[leaf] at (B\k) {$\a$};
}
\coordinate (C0) at (7.40,.82);
\node[leaf] at (C0) {$13$};

\foreach \j in {0,1,2,3}{
  \pgfmathtruncatemacro{\l}{2*\j}\pgfmathtruncatemacro{\r}{2*\j+1}
  \coordinate (A1\j) at ($(A\l)!.5!(A\r)+(0,.58)$);
  \draw[edge] (A1\j) -- ([yshift=.18cm]A\l); \draw[edge] (A1\j) -- ([yshift=.18cm]A\r);
}
\foreach \j in {0,1}{
  \pgfmathtruncatemacro{\l}{2*\j}\pgfmathtruncatemacro{\r}{2*\j+1}
  \coordinate (A2\j) at ($(A1\l)!.5!(A1\r)+(0,.58)$);
  \draw[edge] (A2\j) -- (A1\l); \draw[edge] (A2\j) -- (A1\r);
}
\coordinate (A3) at ($(A20)!.5!(A21)+(0,.58)$);
\draw[edge] (A3) -- (A20); \draw[edge] (A3) -- (A21);
\foreach \j in {0,1}{
  \pgfmathtruncatemacro{\l}{2*\j}\pgfmathtruncatemacro{\r}{2*\j+1}
  \coordinate (B1\j) at ($(B\l)!.5!(B\r)+(0,.58)$);
  \draw[edge] (B1\j) -- ([yshift=.18cm]B\l); \draw[edge] (B1\j) -- ([yshift=.18cm]B\r);
}
\coordinate (B2) at ($(B10)!.5!(B11)+(0,.58)$);
\draw[edge] (B2) -- (B10); \draw[edge] (B2) -- (B11);

\coordinate (J1) at ($(B2)!.5!(C0)+(0,1.16)$);
\draw[jedge] (J1) -- (B2); \draw[jedge] (J1) -- ([yshift=.18cm]C0);
\coordinate (R) at ($(A3)!.5!(J1)+(0,.58)$);
\draw[jedge] (R) -- (A3); \draw[jedge] (R) -- (J1);
\foreach \n in {A10,A11,A12,A13,A20,A21,A3,B10,B11,B2}{\node[tnode] at (\n) {};}
\foreach \n in {J1,R}{\node[jnode] at (\n) {};}

\draw[decorate,decoration={brace,mirror,amplitude=2.5pt},mypool,line width=.4pt]
  (.325,.51) -- (4.415,.51) node[midway,below=2pt,small] {$8$ cells};
\draw[decorate,decoration={brace,mirror,amplitude=2.5pt},mypool,line width=.4pt]
  (4.755,.51) -- (6.765,.51) node[midway,below=2pt,small] {$4$ cells};
\draw[decorate,decoration={brace,mirror,amplitude=2.5pt},mypool,line width=.4pt]
  (7.175,.51) -- (7.625,.51) node[midway,below=2pt,small] {$1$ cell};
\node[small,anchor=south,text=mypool] at ($(R)+(0,.13)$) {root: $13$ cells};
\node[small,anchor=west,text=mypool] at ($(J1)+(.19,.03)$) {$5$ cells};
\draw[mypool,line width=.5pt,{Latex[length=1.3mm]}-{Latex[length=1.3mm]}]
  (8.20,.82) -- (8.20,3.14);
\node[small,rotate=90,anchor=north] at (8.30,1.98) {height $\lceil\log_2 M\rceil=4$};

\draw[panel,draw=mypool,fill=mypool!3] (9.38,2.28) rectangle (14.92,4.32);
\node[title,text=mypool] at (9.60,4.02) {1. Build balanced chunks};
\node[small,anchor=west,align=left] at (9.60,3.45)
  {Median-split mean keys along\\their principal direction.};
\node[small,anchor=west,align=left] at (9.60,2.72)
  {Compute summaries from the cells\\(Eq.~\eqref{eq:fsb-node}).};

\draw[panel,draw=mypool,densely dashed,fill=mypool!3]
  (9.38,.02) rectangle (14.92,2.04);
\node[title,text=mypool] at (9.60,1.75) {2. Join roots, smallest first};
\node[small,anchor=west] at (9.60,1.30)
  {$4+1\rightarrow5,\qquad 8+5\rightarrow13$};
\node[small,anchor=west,align=left] at (9.60,.60)
  {Weighted mean of child centers;\\radius encloses both child balls\\(Eq.~\eqref{eq:fsb-join}).};
\end{tikzpicture}}
\caption{\textbf{Streaming construction of one FS-B forest, per head.}
Solid edges form balanced trees within power-of-two chunks; dashed edges
join their roots, smallest first. Leaf numbers denote arrival order, while
their arrangement follows mean-key splits. The $13=8+4+1$ example has height
$\lceil\log_2 13\rceil=4$. Right: the two construction steps; node summaries
are defined in Eqs.~\eqref{eq:fsb-node}--\eqref{eq:fsb-join}.}
\label{fig:fsb-construction}
\end{figure}

\paragraph{Beam search and reranking.}
Search expands the current beam level by level and keeps the $W^{\mathrm b}$
highest-priority nodes under Eq.~\eqref{eq:fsb-bound}. A leaf reached early
on a short branch keeps its slot, while the remaining slots expand children.
The bound holds for the mean-key score of Eq.~\eqref{eq:qqr-score}, not for
the token-level score of Eq.~\eqref{eq:fsb-rerank} or for a cell's
usefulness to the answer; larger $c$ favors subtrees with dispersed mean
keys. We set $c=8$ on S-NIAH at $341$M and use it unchanged at $743$M. The two candidate sets are merged without a quota for either forest.
In Eq.~\eqref{eq:fsb-rerank}, the log-sum-exp term is the logarithm of a
cell's unnormalized token-attention mass, and the prior affects selection
only. Node summaries never replace token values.

\paragraph{Streaming inference.}
Prefill processes query blocks in fixed-size chunks. During decoding,
the cache keeps only the active chunk trees. A completed larger chunk
replaces its constituent smaller trees, and unchanged chunks are reused. For
$M$ admitted cells, these trees and their joining nodes contain fewer than
$2M$ nodes, giving $O(Md)$ tree storage per head.

\subsection{Cost}
\label{app:fsb-cost}
Let $M\le\lfloor N/s\rfloor=\lfloor N/32\rfloor$ be the number of admitted cells in a forest;
the filtered forest can hold fewer cells than the full-prefix forest. For a
balanced tree and $C_i=|\mathcal B_i|$ candidates, search, token-key
reranking and readout take $O(d[W^{\mathrm b}\log M+sC_i+sK])$ work per
query, with $C_i\le2W^{\mathrm b}$ for the two forests. Fixed budgets
therefore give $O(N\log N)$ routing, reranking and token attention over the
sequence, including the fixed local window.

\paragraph{Construction with radix sorting.}
Assume fixed-width keys for the split projections and stable radix sorting,
so sorting $m$ projections costs $O(m)$ work. Computing the principal
direction, projections and summaries at a node costs $O(md)$, hence a
chunk of $2^k$ cells costs $O(d2^k k)$ over its $k$ split levels.
At most $\lfloor M/2^k\rfloor$ chunks of that size complete, and completed
trees are reused. Summing over chunk sizes gives
\begin{equation}
  \label{eq:fsb-build-cost}
  O\!\left(\sum_{k=1}^{\lfloor\log_2 M\rfloor}
    \frac{M}{2^k}\,d2^k k\right)=O(Md\log^2 M).
\end{equation}
Two forests change only the constant factor. Including search and readout,
the algorithmic sequence cost is $O(N\log^2 N)$ at fixed dimensions and
budgets, still strictly subquadratic. This bound assumes that updates reuse
stored chunks without copying the growing forest at every block.
Comparison sorting at every split would instead give construction work
$O(Md\log^2 M+M\log^3 M)$; the stated bound uses fixed-width radix sorting
or linear-time median partitioning.

\paragraph{Reference streaming overhead.}
The reference cache reuses completed trees but reallocates its growing
admitted-cell index array and repacks all active node tables after updates.
At fixed block size, these copies can accumulate $O(M^2d)$ memory traffic
over a growing decode stream, outside the algorithmic construction bound.
Geometric buffer growth and persistent chunk storage can avoid this copying
without changing tree contents or search. The measured reference throughput
includes these overheads. The token archive uses $O(Nd)$ space and the active
decode trees use $O(Md)$ storage.

\subsection{Forest ablation}
\label{app:fsb-forest-ablation}
Table~\ref{tab:fsb-forest-ablation} compares FS-B's two forests with
single-forest ablations, changing only the inference-time candidate generator;
the checkpoints and the four-cell token read are unchanged. At $800$M,
FS-B matches or exceeds the full-prefix width-$32$ ablation at every length,
by up to $0.8$ mean points at the displayed precision. The width-$16$
full-prefix ablation has the highest mean at $4$k ($37.9$ versus $37.7$
for FS-B). At $340$M, the width-$32$ ablation leads the suite mean at all
four lengths by up to $1.3$ points. Its S-NIAH-2 scores match or exceed
FS-B's; its S-NIAH-3 scores are higher at $2$k--$8$k, but lower at $16$k
($24$ versus $27$). At $800$M, FS-B has higher FWE scores than the
width-$32$ ablation at every length and higher MV-NIAH scores from $4$k onward.

Two width-$16$ beams supply at most $32$ distinct candidates after
deduplication, so a width-$32$ single forest matches the maximum candidate
count. A width-$16$ single forest has a smaller candidate budget and scores
below width $32$ at $8$k--$16$k at both scales. Equal maximum candidate
counts do not imply equal runtime. The norm-filtered width-$16$ ablation
retains $41$--$74\%$ of cells by task, using a
median key-norm threshold calibrated on eight $4096$-token FineWeb-Edu
batches. Relative to FS-B, its suite mean is lower by $1.5$--$2.9$ points
at $340$M and $0.2$--$1.5$ points at $800$M.

\begin{table}[ht]
\caption{\textbf{Forest ablation of FS-B} ($0$--$100$, RULER protocol of Table~\ref{tab:ruler}).
All arms run on the same checkpoints with the same four-cell token read; only the
inference-time candidate generator differs: FS-B's two forests as rendered in
Table~\ref{tab:ruler}, the full-prefix forest alone with a width-$32$ or
width-$16$ beam, or the norm-filtered forest alone with a width-$16$ beam.
Bold marks the best of the four arms per column.}
\label{tab:fsb-forest-ablation}
\begin{center}
\footnotesize\setlength{\tabcolsep}{2.5pt}
\resizebox{\textwidth}{!}{\begin{tabular}{@{}l cccc cccc cccc cccc@{}}
\toprule
 & \multicolumn{4}{c}{FS-B (two forests)} & \multicolumn{4}{c}{full, $W^{\mathrm b}{=}32$} & \multicolumn{4}{c}{filtered, $W^{\mathrm b}{=}16$} & \multicolumn{4}{c}{full, $W^{\mathrm b}{=}16$} \\
\cmidrule(lr){2-5}\cmidrule(lr){6-9}\cmidrule(lr){10-13}\cmidrule(lr){14-17}
Task & $2$k & $4$k & $8$k & $16$k & $2$k & $4$k & $8$k & $16$k & $2$k & $4$k & $8$k & $16$k & $2$k & $4$k & $8$k & $16$k \\
\midrule
\multicolumn{17}{@{}l}{\emph{$\approx$340M scale}} \\
\midrule
S-NIAH-1 & $\mathbf{100}$ & $\mathbf{99}$ & $\mathbf{99}$ & $\mathbf{99}$ & $\mathbf{100}$ & $\mathbf{99}$ & $\mathbf{99}$ & $\mathbf{99}$ & $99$ & $\mathbf{99}$ & $\mathbf{99}$ & $\mathbf{99}$ & $\mathbf{100}$ & $\mathbf{99}$ & $\mathbf{99}$ & $\mathbf{99}$ \\
S-NIAH-2 & $\mathbf{100}$ & $99$ & $89$ & $60$ & $\mathbf{100}$ & $\mathbf{100}$ & $\mathbf{92}$ & $\mathbf{67}$ & $97$ & $99$ & $68$ & $49$ & $\mathbf{100}$ & $99$ & $77$ & $44$ \\
S-NIAH-3 & $91$ & $78$ & $64$ & $\mathbf{27}$ & $\mathbf{95}$ & $\mathbf{84}$ & $\mathbf{71}$ & $24$ & $75$ & $61$ & $51$ & $22$ & $90$ & $79$ & $70$ & $26$ \\
MK-NIAH-1 & $27$ & $25$ & $19$ & $\mathbf{24}$ & $27$ & $\mathbf{28}$ & $\mathbf{23}$ & $21$ & $29$ & $24$ & $22$ & $19$ & $\mathbf{30}$ & $20$ & $\mathbf{23}$ & $19$ \\
MK-NIAH-2 & $\mathbf{2}$ & $\mathbf{0}$ & $\mathbf{0}$ & $\mathbf{1}$ & $\mathbf{2}$ & $\mathbf{0}$ & $\mathbf{0}$ & $\mathbf{1}$ & $\mathbf{2}$ & $\mathbf{0}$ & $\mathbf{0}$ & $\mathbf{1}$ & $\mathbf{2}$ & $\mathbf{0}$ & $\mathbf{0}$ & $0$ \\
MK-NIAH-3 & $\mathbf{2}$ & $0$ & $\mathbf{0}$ & $\mathbf{0}$ & $1$ & $0$ & $\mathbf{0}$ & $\mathbf{0}$ & $1$ & $\mathbf{1}$ & $\mathbf{0}$ & $\mathbf{0}$ & $1$ & $0$ & $\mathbf{0}$ & $\mathbf{0}$ \\
MQ-NIAH & $23.8$ & $\mathbf{22.5}$ & $18.5$ & $12.5$ & $23.8$ & $22.2$ & $\mathbf{19.0}$ & $\mathbf{15.5}$ & $22.8$ & $21.8$ & $16.8$ & $10.2$ & $\mathbf{24.8}$ & $21.5$ & $18.5$ & $14.0$ \\
MV-NIAH & $\mathbf{24.5}$ & $22.0$ & $23.5$ & $\mathbf{21.0}$ & $\mathbf{24.5}$ & $22.5$ & $\mathbf{25.5}$ & $\mathbf{21.0}$ & $23.0$ & $21.5$ & $23.0$ & $16.0$ & $24.2$ & $\mathbf{23.2}$ & $22.0$ & $18.0$ \\
VT & $\mathbf{3.2}$ & $1.2$ & $1.0$ & $1.0$ & $3.0$ & $\mathbf{1.4}$ & $\mathbf{1.6}$ & $1.2$ & $2.8$ & $1.2$ & $1.0$ & $1.2$ & $3.0$ & $1.2$ & $1.0$ & $\mathbf{1.4}$ \\
CWE & $5.2$ & $10.3$ & $6.5$ & $1.8$ & $\mathbf{6.7}$ & $10.0$ & $6.6$ & $\mathbf{2.0}$ & $5.0$ & $\mathbf{10.6}$ & $6.6$ & $1.8$ & $5.3$ & $10.4$ & $\mathbf{7.3}$ & $1.3$ \\
FWE & $9.0$ & $\mathbf{6.0}$ & $\mathbf{11.7}$ & $5.0$ & $9.3$ & $4.7$ & $9.0$ & $5.7$ & $8.7$ & $3.7$ & $6.7$ & $3.0$ & $\mathbf{9.7}$ & $\mathbf{6.0}$ & $9.7$ & $\mathbf{6.7}$ \\
QA-1 & $\mathbf{23}$ & $10$ & $4$ & $4$ & $22$ & $9$ & $\mathbf{6}$ & $\mathbf{5}$ & $18$ & $\mathbf{12}$ & $\mathbf{6}$ & $3$ & $\mathbf{23}$ & $10$ & $3$ & $4$ \\
QA-2 & -- & -- & $9$ & $6$ & -- & -- & $\mathbf{10}$ & $6$ & -- & -- & $8$ & $\mathbf{7}$ & -- & -- & $7$ & $6$ \\
\midrule
Available-task mean & $34.2$ & $31.1$ & $26.6$ & $20.2$ & $\mathbf{34.5}$ & $\mathbf{31.7}$ & $\mathbf{27.9}$ & $\mathbf{20.6}$ & $31.9$ & $29.6$ & $23.7$ & $17.9$ & $34.4$ & $30.8$ & $26.0$ & $18.4$ \\
\midrule
\multicolumn{17}{@{}l}{\emph{$\approx$800M scale}} \\
\midrule
S-NIAH-1 & $\mathbf{97}$ & $98$ & $97$ & $\mathbf{100}$ & $95$ & $98$ & $99$ & $99$ & $\mathbf{97}$ & $\mathbf{99}$ & $\mathbf{100}$ & $99$ & $\mathbf{97}$ & $\mathbf{99}$ & $98$ & $\mathbf{100}$ \\
S-NIAH-2 & $\mathbf{100}$ & $\mathbf{100}$ & $\mathbf{95}$ & $88$ & $\mathbf{100}$ & $\mathbf{100}$ & $\mathbf{95}$ & $\mathbf{90}$ & $\mathbf{100}$ & $\mathbf{100}$ & $\mathbf{95}$ & $81$ & $98$ & $\mathbf{100}$ & $87$ & $80$ \\
S-NIAH-3 & $\mathbf{96}$ & $93$ & $87$ & $84$ & $\mathbf{96}$ & $93$ & $87$ & $\mathbf{87}$ & $88$ & $88$ & $84$ & $81$ & $93$ & $\mathbf{94}$ & $\mathbf{90}$ & $82$ \\
MK-NIAH-1 & $50$ & $\mathbf{34}$ & $28$ & $\mathbf{31}$ & $46$ & $25$ & $\mathbf{30}$ & $29$ & $\mathbf{52}$ & $28$ & $28$ & $30$ & $48$ & $28$ & $27$ & $26$ \\
MK-NIAH-2 & $9$ & $\mathbf{4}$ & $1$ & $2$ & $5$ & $\mathbf{4}$ & $1$ & $2$ & $8$ & $3$ & $1$ & $\mathbf{3}$ & $\mathbf{11}$ & $1$ & $\mathbf{2}$ & $2$ \\
MK-NIAH-3 & $\mathbf{2}$ & $\mathbf{1}$ & $0$ & $\mathbf{0}$ & $\mathbf{2}$ & $0$ & $\mathbf{1}$ & $\mathbf{0}$ & $1$ & $0$ & $0$ & $\mathbf{0}$ & $\mathbf{2}$ & $0$ & $0$ & $\mathbf{0}$ \\
MQ-NIAH & $\mathbf{45.8}$ & $28.3$ & $28.8$ & $27.3$ & $45.0$ & $\mathbf{30.2}$ & $27.2$ & $\mathbf{27.8}$ & $40.5$ & $29.2$ & $\mathbf{29.0}$ & $25.8$ & $38.2$ & $29.0$ & $25.0$ & $25.5$ \\
MV-NIAH & $38.3$ & $\mathbf{30.3}$ & $\mathbf{29.8}$ & $\mathbf{26.3}$ & $\mathbf{39.5}$ & $28.8$ & $26.2$ & $24.5$ & $38.0$ & $29.8$ & $27.2$ & $26.2$ & $34.2$ & $28.2$ & $26.2$ & $24.8$ \\
VT & $3.8$ & $1.4$ & $\mathbf{1.0}$ & $\mathbf{0.6}$ & $4.0$ & $2.0$ & $0.8$ & $0.4$ & $\mathbf{4.8}$ & $2.0$ & $0.8$ & $0.2$ & $4.0$ & $\mathbf{2.4}$ & $\mathbf{1.0}$ & $\mathbf{0.6}$ \\
CWE & $7.8$ & $26.4$ & $8.9$ & $\mathbf{4.4}$ & $7.8$ & $29.2$ & $8.1$ & $3.4$ & $\mathbf{10.0}$ & $24.9$ & $\mathbf{9.9}$ & $3.2$ & $9.6$ & $\mathbf{30.1}$ & $7.4$ & $2.3$ \\
FWE & $17.7$ & $8.7$ & $7.0$ & $8.0$ & $15.3$ & $7.3$ & $3.3$ & $5.7$ & $10.3$ & $4.7$ & $\mathbf{7.7}$ & $\mathbf{15.0}$ & $\mathbf{23.0}$ & $\mathbf{10.7}$ & $7.3$ & $8.3$ \\
QA-1 & $38$ & $27$ & $13$ & $10$ & $\mathbf{40}$ & $29$ & $\mathbf{14}$ & $\mathbf{13}$ & $39$ & $26$ & $13$ & $11$ & $\mathbf{40}$ & $\mathbf{32}$ & $\mathbf{14}$ & $12$ \\
QA-2 & -- & -- & $17$ & $\mathbf{14}$ & -- & -- & $\mathbf{18}$ & $\mathbf{14}$ & -- & -- & $15$ & $9$ & -- & -- & $14$ & $11$ \\
\midrule
Available-task mean & $\mathbf{42.1}$ & $37.7$ & $\mathbf{31.8}$ & $\mathbf{30.4}$ & $41.3$ & $37.2$ & $31.6$ & $\mathbf{30.4}$ & $40.7$ & $36.2$ & $31.6$ & $29.6$ & $41.5$ & $\mathbf{37.9}$ & $30.7$ & $28.8$ \\
\bottomrule
\end{tabular}}
\end{center}
\end{table}

\subsection{Retrieval-weighted CPT}
\label{app:fsb-cpt}
\paragraph{Recall masks.}
Let $f_t$ indicate a recall start or continuation, identified by a unique
earlier four-token match more than $512$ tokens away. A start matches the
span beginning at the target token with a different preceding token; a
continuation matches the preceding four tokens and their next target. Let
$\chi_t$ mark a recall start within $64$ tokens of the previous start whose
source is more than $64$ tokens from the previous source, which
distinguishes switching to another span from returning to the same one. The
masks respect document boundaries and do not depend on the model's
confidence. Only natural FineWeb-Edu text is used; benchmark prompts and
answers are never training examples or sources of loss weights.

\paragraph{Schedule.}
Starting from the base $800$M FS checkpoint, CPT has two $800$-step stages
of $262{,}144$ tokens per step with $\beta=9$; $\beta_{\mathrm{sw}}$ is $9$ in
the first stage and $49$ in the second, so a switch token has weight $19$ and
then $59$. In each stage the learning rate warms up for $20$ steps to
$4\times10^{-5}$ and decays to $4\times10^{-6}$. The total is $1600$ steps
and $0.42$B tokens, all at $4096$ tokens per sequence, so evaluation beyond
$4$k remains zero-shot extrapolation. During CPT, FS scores the eligible
prefix and reads $16$ cells per head; at evaluation, FS-B searches the two
forests and reads four. The forests are not in the training loop, and no
routing parameters are learned.

\paragraph{CPT ablation with single-forest readout.}
Table~\ref{tab:fsb-cpt-eval} compares the base and CPT checkpoints under
the single-forest ablation read paths, holding each selector fixed: the
$0.42$B-token continuation improves every arm and length, with the largest
gains at $16$k. The gains concentrate on the retrieval tasks that the
weighting targets (S-NIAH-1/3 improve by $4$--$18$ points and multi-key
and multi-value retrieval by $4$--$8$), with small regressions on CWE of
up to $7.8$ points and on FWE of up to $3.7$. Reading the same CPT
checkpoint with FS-B's two-forest search gives $42.1/37.7/31.8/30.4$
(Table~\ref{tab:ruler}), exceeding the full-prefix width-$32$ ablation at
$2$k--$8$k and matching it at $16$k at the displayed precision.
The per-task breakdown of the CPT arms is in
Table~\ref{tab:fsb-forest-ablation}. The pools store aggregate scores per
task, so no prompt-level intervals are reported.

\begin{table}[ht]
\caption{\textbf{Recall-weighted CPT under single-forest ablations}
($0$--$100$; RULER available-task means over $12$ tasks at $2$k/$4$k and
$13$ at $8$k/$16$k, protocol of Table~\ref{tab:ruler}). All rows start from
the base $743$M FS checkpoint; CPT rows are the Table~\ref{tab:ruler} FS-B
checkpoint evaluated with a single-forest ablation. FS-B denotes the
two-forest design, whose results appear in Table~\ref{tab:ruler}.}
\label{tab:fsb-cpt-eval}
\centering\footnotesize\setlength{\tabcolsep}{4pt}
\begin{tabular}{@{}ll*{5}{r}@{}}
\toprule
Read path & Weights & $2$k & $4$k & $8$k & $16$k & Mean \\
\midrule
Full-prefix forest (W32) & base FS & $40.9$ & $36.0$ & $30.4$ & $28.2$ & $33.9$ \\
& $+$weighted CPT & $41.3$ & $37.2$ & $31.6$ & $30.4$ & $35.1$ \\
Norm-filtered forest (W16) & base FS & $39.6$ & $35.1$ & $29.8$ & $27.5$ & $33.0$ \\
& $+$weighted CPT & $40.7$ & $36.2$ & $31.6$ & $29.6$ & $34.5$ \\
\bottomrule
\end{tabular}
\end{table}

\subsection{Relation to chunk and hierarchical retrieval}
\label{app:retrieval-comparison}
Several methods retrieve distant chunks. HCAM~\citep{lampinen2021hcam} and
the Hierarchical Sparse Attention of RAMba~\citep{hu2025hsa} score chunk
summaries, attend within selected chunks, and combine the chunk outputs; DRT's
Grouped Cross-Attention~\citep{hu2025gca} retrieves past chunks for each next
chunk. Their routing scores weight the chunk outputs, which gives the router a
gradient, and DRT and RAMba offload growing token memory to the CPU while
keeping chunk keys on the GPU. FreeSolo instead uses hard cell selection
followed by one softmax over the selected tokens, so its shared projections
learn through the token readout and local attention; FS-L bounds the accessible
cells by $P+W$ per head, and FS-B searches two content forests that together cover every cell.
HMT~\citep{he2025hmt} recalls compressed segment memories within a recurrent
architecture, whereas FreeSolo keeps the token-level keys and values of every
retrievable cell. Treeformer's TF-attention~\citep{madaan2023treeformer} learns decision-tree
partitions and attends to keys routed to the query's leaf; FS-B builds balanced
partitions from the mean keys of causally available cells and searches several
branches with query-dependent priorities. The published protocols also differ:
DRT uses task fine-tuning and RAMba's RULER tests include synthetic fine-tuning
at $4$k, whereas our models use next-token pretraining only, so we do not
compare published scores.

\end{document}